\documentclass{article}
\usepackage{xcolor}

\newif\ifSWBfinal
\SWBfinaltrue                  
\ifdefined\SWBFINAL\SWBfinaltrue\fi

\newif\ifSWBdata
\SWBdatatrue                   
\ifdefined\SWBDATA\SWBdatatrue\fi

\newif\ifSWBcameraready
\SWBcamerareadytrue
\ifdefined\SWBCAMERAREADY\SWBcamerareadyfalse\fi

\definecolor{SWBph}{rgb}{0.85,0.05,0.05}

\newcommand{\CensusDOI}{\href{https://doi.org/10.5281/zenodo.22773066}{10.5281/zenodo.22773066}}

\ifSWBcameraready
  \newcommand{\CensusURL}{\CensusDOI}
  \newcommand{\CensusStatus}{is released, versioned and archived}
  \newcommand{\CensusAbstract}{\noindent Code: \CensusURL{}\\
    Data: \CensusDOI{}}
  
  \newcommand{\CensusFullLine}{}
\fi

\ifSWBcameraready
  \usepackage[final,dblblindworkshop]{neurips_2026}
\else
  \usepackage[dblblindworkshop]{neurips_2026}   
\fi
\workshoptitle{Foundation Models for Temporal Systems: From Forecasting to World Modeling}
\usepackage[utf8]{inputenc}
\usepackage[T1]{fontenc}
\usepackage{xcolor}
\usepackage{hyperref}
\definecolor{swblink}{RGB}{16,66,138}    
\definecolor{swbcite}{RGB}{17,102,86}    
\definecolor{swburl}{RGB}{38,79,156}     
\hypersetup{colorlinks=true,
            linkcolor=swblink,
            citecolor=swbcite,
            urlcolor=swburl,
            filecolor=swburl,
            linktoc=all}
\ifSWBcameraready
  \hypersetup{pdftitle={The Social World Model Evaluation Gap},
              pdfauthor={Denglin Jiang and Annie Wang},
              pdfsubject={NeurIPS 2026 Workshop on Foundation Models for Temporal Systems (FMTS)}}
\fi
\usepackage{url}
\ifSWBcameraready

\else
  \usepackage{xurl}   
\fi
\usepackage{booktabs}
\usepackage{array}       
\usepackage{tabularx}    
\usepackage{longtable}   
\usepackage{graphicx}
\usepackage{amsmath}
\usepackage{amssymb}     
\usepackage{xcolor}      
\usepackage{needspace}   
\newcolumntype{Y}{>{\raggedright\arraybackslash}X}
\newcolumntype{P}[1]{>{\raggedright\arraybackslash}p{#1}}

\newcommand{\NMetRejectedPassThree}{8}

\newcommand{\NRows}{123}
\newcommand{\NRowsExcluded}{10}
\newcommand{\NMethods}{113}
\newcommand{\NEgo}{54}

\newcommand{\NEgoHuman}{24}

\newcommand{\NInterventional}{9}

\newcommand{\NIntervFoundBySweep}{8}

\newcommand{\NIntervOriginalScreen}{1}

\newcommand{\NRegimeCoded}{106}
\newcommand{\NRegimeUncoded}{7}

\newcommand{\NRegimeInterv}{9}

\newcommand{\NRegimeStatic}{25}

\newcommand{\NRegimeTask}{40}

\newcommand{\NRegimeObs}{32}

\newcommand{\NRegimeCompound}{5}

\newcommand{\RegimeCompoundList}{\citet{kwon2020risk} (\texttt{behavior fidelity/intervention}); \citet{nikolaidis2017mutual} (\texttt{goal completion/intervention}); \citet{valevski2024gamengen} (\texttt{rollout error/human judgment}); \citet{wang2026metamind} (\texttt{goal completion/novel partner}); \citet{zhou2025socialworldmodels} (\texttt{static QA/goal completion})}
\newcommand{\NRandomized}{3}

\newcommand{\NRegimeCodedNoPhys}{91}

\newcommand{\NIntervNoPhys}{9}

\newcommand{\NIntervForecastDirect}{2}

\newcommand{\NIntervForecastPartial}{5}

\newcommand{\NIntervForecastNone}{2}

\newcommand{\NIntervForecastDirectOffline}{2}

\newcommand{\NIntervDirectOrOutcome}{3}

\newcommand{\NDirectIfRejectedAdmitted}{2}

\newcommand{\NDirectNoPhys}{2}

\newcommand{\NBoundaryChecked}{32}

\newcommand{\NBoundaryPhysTask}{8}

\newcommand{\NBoundaryNoForecast}{19}

\newcommand{\NBoundaryYesForecast}{13}

\newcommand{\NRegimeCodedNoBoundary}{87}

\newcommand{\NStrictSubset}{72}
\newcommand{\NStrictSubsetDirect}{2}
\newcommand{\NDirectFoundBySweep}{2}

\newcommand{\NDirectOriginalScreen}{0}

\newcommand{\IntervDirectCites}{\citet{mandel2014ope} and \citet{soh2018trusttransfer}}
\newcommand{\NFamLLMChecked}{25}

\newcommand{\NFamLLMAnyQualifying}{0}

\newcommand{\NFamLLMHumanStudy}{3}

\newcommand{\NFamLLMAssignedHuman}{1}

\newcommand{\NFamLLMForecastVsHuman}{1}

\newcommand{\NSecondStudies}{9}
\newcommand{\NSecondAgree}{7}

\newcommand{\NSecondLLMCoded}{98}
\newcommand{\NSecondLLMAgree}{97}
\newcommand{\NSecondChanged}{0}

\newcommand{\NSourceVersionRecorded}{4}
\newcommand{\NSourceVersionMissing}{21}
\newcommand{\NSourceVersionPreprint}{15}
\newcommand{\NSourceVersionProceedings}{6}

\newcommand{\NIntervOneLimb}{32}

\newcommand{\NIntervIfRejectedAdmitted}{17}

\newcommand{\NScorableIfRejectedAdmitted}{114}

\newcommand{\NEgoAlt}{63}

\newcommand{\NInterventionalAlt}{9}
\newcommand{\YearFirst}{1999}
\newcommand{\YearLast}{2026}

\newcommand{\NFamPhys}{19}

\newcommand{\NFamPhysScored}{15}

\newcommand{\NSweeps}{4}

\title{The Social World Model Evaluation Gap}
\ifSWBcameraready
  \author{Denglin Jiang \\
    New York University \\
    \texttt{florence.jiang@nyu.edu} \And
    Annie Wang \\
    New York University \\
    \texttt{yw4509@nyu.edu}}
\else
  \author{Anonymous}
\fi

\begin{document}
\maketitle
\suppressfloats[t]
\begin{abstract}
A negotiation assistant can reach a good deal without accurately predicting the
other person's response. We ask whether evaluations of social world models test
that prediction directly. We audit \NMethods{} works published from
\YearFirst{} to \YearLast{}. \NInterventional{} primary evaluations measure a
real person's response to an action selected by the system or experimenter.
Only \NIntervForecastDirect{} also compare an explicit response forecast with
that response. Their predictors use information available when the action is
chosen, and the evaluated responses were excluded from fitting; both randomize
the intervention. A separate full-text check of \NFamLLMChecked{}
language-model works finds no experiment meeting these conditions. The other
evaluations test current-state inference, task success, or prediction from
recorded actions. We propose a 1-step negotiation experiment that separates
these questions. The findings describe the audited sample, not the prevalence
of this evaluation gap across the field.
\\[2pt]
\ifSWBcameraready
\noindent Code and data: \CensusDOI{}.
\else
\CensusAbstract{}
\fi

\end{abstract}

\section{Introduction}
\label{sec:intro}

Imagine an assistant negotiating a deal for you. It can make a concession or
repeat the offer with a justification. A model used to compare these moves
should predict how the other person, the \emph{counterpart}, will respond to
each. We call the component that makes this prediction a \emph{social world
model} \citep{zhou2025socialworldmodels, yu2026socialworldmodels,
tsvilodub-etal-2026-emergent}.

Suppose the assistant gets a good deal. That establishes task success, but it
does not show that its predictions were accurate. Recognizing that the other
person is frustrated is different again: it estimates a current state, rather
than predicting a response to a proposed move. These distinctions matter when
a model is used to simulate an interaction \citep{zhang2026digitaltwin} or
choose what to do next.

This is a temporal forecasting problem. Each move changes the context for the
next response. We study its simplest form: 1 system action followed by 1 human
response. Recorded conversations alone may not settle it. People may concede
when they already expect agreement, so a model can learn that concessions
precede agreement without learning whether a concession would help in a
different situation. The issue is the difference between observing an action
and assigning it \citep{pearl2009causality, robins1986new}.

We ask whether a model's prediction is checked against a real person's response
to an action the system assigned. Only \NIntervForecastDirect{} primary
evaluations in our sample meet this test. We checked every experiment in the
\NFamLLMChecked{} large language model (LLM) works and found none there. We
also describe how to run the test for a negotiation assistant. This is evidence
about evaluation practice in the audited sample, not a claim that the
capability is absent.

\Needspace*{7\baselineskip}
\section{What Would Test a Response Forecast?}
\label{sec:question}

We call an experiment a \emph{direct forecast evaluation} when it meets 3
conditions:
\begin{enumerate}\setlength{\itemsep}{0pt}\setlength{\topsep}{1pt}\setlength{\parskip}{0pt}
\item \textbf{An assigned action.} The system or experimenter selects the
  action. It is not simply an action that a person chose in a recorded
  conversation.
\item \textbf{A human response.} A real person responds. Human ratings of a
  conversation between 2 models do not meet this condition.
\item \textbf{A scored forecast.} The model uses the candidate action to
  predict that response, and the prediction is compared with the measured
  response.
\end{enumerate}
All 3 must hold in the same experiment. A study meeting the first 2 can
measure a human response without testing a forecast. We call this broader
category an \emph{assigned-action human-response study}. Randomized assignment
is an additional property, not a requirement for every direct evaluation.

\paragraph{What information may the model use?}
The forecast may use the history available when the action is chosen and the
candidate action, or a fully specified assigned intervention. A response cannot
be used to train or update the predictor before that response is scored. This
allows testing on held-out responses, and it allows online evaluation provided
each prediction is scored before the observed response updates the model.
Neither design by itself establishes prospective performance when the system
chooses its own actions. A robot demonstration's programmed success or failure
can be part of the assigned intervention; an outcome learned only after acting
is different, and a forecast using it is \emph{outcome-conditioned} and receives
a partial grade. Unless we say otherwise, every count below is of primary
evaluations (\S\ref{sec:method}); only the LLM subset is checked across all its
experiments.

\paragraph{Why distinguish observed and assigned actions?}
Recorded conversations show what followed moves that people chose. To compare
the consequences of different moves, we instead ask what response would follow
if the system selected each move in the same situation. These questions can
have different answers: people may concede when they already expect agreement,
so a model can learn that concessions precede agreement without learning
whether a concession would help elsewhere. Randomization addresses this source
of confounding. Formally, with $h$ the interaction history, $a$ the next move
and $Y$ the person's response, recorded moves test $P(Y\mid h,a)$ while
choosing a move calls for $P(Y\mid h,\operatorname{do}(a))$, where
$\operatorname{do}(a)$ means setting the move to $a$. Observational data
support the second comparison only under further assumptions: the history
captures the relevant confounders, the move is well defined, and the data cover
it \citep{robins1986new, pearl2009causality}. A fixed policy can still test
forecasts for moves it delivers, but cannot compare them with moves it never
takes.

\begin{table}[htbp]
\centering
\small
\setlength{\tabcolsep}{4pt}
\begin{tabularx}{\linewidth}{@{}P{0.27\linewidth}Yr@{}}
\toprule
Evaluation regime & Question answered & Works \\
\midrule
Current-state inference & What does the other agent think or feel now? & \NRegimeStatic{} \\
Task success & Does the system achieve its objective? & \NRegimeTask{} \\
Observed-action prediction & What follows actions in recorded data? & \NRegimeObs{} \\
Assigned-action human response & How did a person respond to an action the system or procedure selected? & \NRegimeInterv{} \\
\midrule
Total & & \NRegimeCoded{} \\
\bottomrule
\end{tabularx}
\caption{The primary evaluations of the \NRegimeCoded{} scored works. The \NRegimeUncoded{} unscored works are excluded. Of the \NRegimeInterv{} studies in the last row, \NIntervForecastDirect{} score eligible forecasts; \NRandomized{} randomize assignment, including both direct cases.}
\label{tab:regimes}
\end{table}

\section{How We Coded the Corpus}
\label{sec:method}
\label{sec:method:screening}

We classified \NMethods{} works, with 1 database row per work. Of these,
\NRegimeCoded{} report an evaluation we can classify; the remaining
\NRegimeUncoded{} are outside Table~\ref{tab:regimes}. The database also holds
\NRowsExcluded{} benchmark or prior-survey records outside that set, for
\NRows{} rows in all. We searched English-language work from
\YearFirst--\YearLast{} in 3 stages, dated in Appendix~\ref{app:audit}:
\NSweeps{} topic sweeps, the nearest concurrent survey's references
\citep{zhang2026swmsurvey}, and a targeted search for studies that might meet
our criterion.

We code each work's \emph{primary evaluation}: the setting its abstract and
main tables emphasize, called the \emph{headline evaluation} in the codebook.
Secondary studies and ablations do not change that code. For the LLM subset
only, we additionally check every experiment in the full text and appendices.
Appendix~\ref{app:audit} gives the tie-breaking rules and a worked example.

The written admission rule required a scored prediction of another agent, or a
physical world-model foundation. The assembled sample was broader. Of
\NRegimeTask{} task-success works, \NBoundaryPhysTask{} are physical
foundations; the remaining \NBoundaryChecked{} include
\NBoundaryYesForecast{} with a scored prediction of another agent and
\NBoundaryNoForecast{} without one. The latter contain an unscored counterpart
model or one implicit in the policy. \emph{After assembling the sample}, we
broadened the written rule to cover these works. We did not add or remove any
paper when making that change, and the codebook dates it. Pure benchmarks and
agents with no counterpart model remain excluded.

LLMs proposed the initial classifications from primary sources. 1 author then
reviewed every entry and decided which candidates from the targeted search to
include, using quotations from the source papers. That author knew the study's
hypothesis. A second author then classified the decisive cases independently and
blind to those codes: condition (3) for the \NInterventional{} assigned-action
studies, and the 4 flags for all \NFamLLMChecked{} language-model works. That reader agreed
on \NSecondAgree{} of the \NSecondStudies{} grades and \NSecondLLMAgree{} of the
\NSecondLLMCoded{} cells they filled, named the same \NIntervForecastDirect{}
direct evaluations, and found no qualifying language-model experiment; an author
resolved the rest against the source and \NSecondChanged{} codes changed.
Scripts reproduce the counts from
the recorded classifications, and the corpus, codebook, screening record,
per-study evidence and those scripts are
\ifSWBcameraready
released and archived (Appendix~\ref{app:audit}); the archive regenerates
every count in this paper.
\else
archived as an \CensusURL{}, which
regenerates every count in this paper (Appendix~\ref{app:audit}).
\fi

\section{What the Evaluations Establish}
\label{sec:results}
\label{sec:robustness}

Table~\ref{tab:regimes} groups the \NRegimeCoded{} scored works by the question
their primary evaluation answers. Of all \NMethods{} coded works, \NEgo{}
condition on the system's action, and \NEgoHuman{} of those use human data or
ratings. The \NInterventional{} assigned-action human-response studies come
from human--robot interaction (HRI) or tutoring, not LLMs. \textbf{Only
\NIntervForecastDirect{} directly score eligible forecasts; both randomize
assignment} (Table~\ref{tab:direct}). The others comprise
\NIntervForecastPartial{} partial evaluations and \NIntervForecastNone{} that
measure responses without scoring forecasts.

\begin{table}[htbp]
\centering
\small
\setlength{\tabcolsep}{4pt}
\begin{tabularx}{\linewidth}{@{}P{0.17\linewidth}P{0.39\linewidth}Y@{}}
\toprule
Study & Assigned intervention and response & Forecast test \\
\midrule
\citet{mandel2014ope} & Next tutoring action, drawn uniformly at random from the valid choices. \emph{Target:} The learner's next observations. & Cross-validated on held-out learner logs. Log-likelihood (Fig. 3). The fold scheme for Fig.~3 is not stated. \\
\citet{soh2018trusttransfer} & A randomly assigned combination of robot demonstration task and programmed success or failure. \emph{Target:} The participant's reported trust on other tasks. & Held-out participants (10 folds) or tasks (12 folds). NLL and mean absolute error (Tables 2-3). \\
\bottomrule
\end{tabularx}
\caption{The \NIntervForecastDirect{} direct evaluations, both randomized. Log-likelihood measures how well a model assigns probability to the observed responses. NLL is negative log-likelihood; MAE is mean absolute error. Full evidence is in Table~\ref{tab:allstudies}.}
\label{tab:direct}
\end{table}

Partial evidence includes trend tests, fits scored on training data, and scores
for a different predictor. \citet{bhat2022clustering} predicts trust using the
outcome already observed at the same task site, an input that would not be
available before acting; the paper also does not state clearly how its reported
prediction error was evaluated. \citet{carroll2019overcooked} does score
human-action predictions on held-out data, but its predictor uses the game
state alone. It therefore does not test whether the human-response forecast
depends on the system's proposed move, as condition~(3) requires.

\paragraph{The LLM subset.}
We checked every experiment in the \NFamLLMChecked{} LLM papers and found none
that met all 3 conditions. \NFamLLMHumanStudy{} include human interaction or
role-play: \citet{qiu2023minddial} reports game outcomes and ratings without
scoring a belief forecast in its human study; \citet{jin2024bipalm} scores
retrospective goal and belief inference; and \citet{ys2026sotopiatom} includes
human role-play as a reference and scores the interaction outcomes, rather than
a model's forecast of the human response. DialToM \citep{yadav2026dialtom} forecasts from
recorded dialogue without assigning a system move to a human counterpart.
SOTOPIA-$\pi$ \citep{wang-etal-2024-sotopia} uses human raters of model--model
conversations. Appendix~\ref{app:llm} separates these forms of human
involvement.

\paragraph{Sensitivity to audit choices.}
The \NIntervForecastDirect{} direct cases survive every exclusion we tried:
dropping the physical foundations, the \NBoundaryNoForecast{} works without a
scored counterpart prediction, or both leaves the count unchanged, as does
admitting the \NMetRejectedPassThree{} disputed candidates under their existing
grades. Appendix~\ref{app:sensitivity} gives each denominator. These checks
cannot tell us how many relevant studies the search missed: both direct cases
were found only in the later, targeted search.

\section{An Experiment That Would Test It}
\label{sec:protocol}

We adapt the \NIntervForecastDirect{} positive precedents
(\IntervDirectCites{}) to a negotiation between a system and 1 person. This is
a proposed study, not a completed experiment.
\begin{enumerate}\setlength{\itemsep}{0pt}\setlength{\topsep}{1pt}\setlength{\parskip}{0pt}
\item \textbf{Define the decision.} Use several negotiation situations and
  select 1 turn per participant under a rule fixed in advance. Record the
  conversation available at that turn as $h$. Specify 2 complete messages: a
  concession of 1 item, and a restatement of the offer with a justification.
  Disclose both at consent.
\item \textbf{Prepare 3 predictors and log every forecast.} Fit a history-only
  predictor, a history-plus-action predictor sharing its training data and model
  class, and an action-only baseline that predicts each move's average
  acceptance rate. Fit all 3 on training participants who are separate from the
  test participants, then freeze them. Before any move is selected, log the
  history-only forecast $\hat p_0=\hat P(Y=1\mid h)$, the action-only forecast,
  and the history-plus-action forecast $\hat p_a=\hat P(Y=1\mid h,a)$ for both
  candidate moves.
\item \textbf{Assign a move and observe.} Choose either message with
  probability $1/2$, independently of the forecasts and participant, and deliver
  it verbatim. Code next-turn acceptance as $Y=1$; an observed nonaccepting
  reply or exit as $Y=0$. Keep lost or unrecorded responses separate as missing,
  and define in advance how ambiguous replies are coded. Record the accepted
  offer's next-turn payoff, or 0 otherwise; this is not eventual negotiation
  value.
\item \textbf{Compare prediction errors against both baselines.} Score each
  predictor on the same observed response with the Brier score
  $(\hat p-Y)^2$, the squared difference between the predicted acceptance
  probability and the 0-or-1 outcome, where lower is better
  \citep{brier1950verification, gneiting2007strictly}. For each baseline,
  subtract the full model's error from the baseline's error on the same
  response and average over participants; a positive mean favors the full
  model. Both comparisons are primary: prespecify them, and require improvement
  over both, with the lower endpoint of each paired 95\% participant-bootstrap
  interval above 0. Report per-move scores, acceptance and payoff as secondary
  outcomes, and missing-response rates and bounds; Appendix~\ref{app:protocol}
  fixes the sample size for a stated interval width.
\end{enumerate}
Beating the history-only predictor shows that knowing the move helps; beating
the action-only predictor shows that the conversation helps beyond each move's
average acceptance rate. Without the second comparison, an apparent gain could
reflect only the average difference between the 2 moves. Neither reveals both
responses 1 person could have given, nor establishes long-term decision value:
in \citet{mandel2014ope}, held-out likelihood improved while the selected
policy's reward fell.

\paragraph{Limitations.}
\label{sec:limitations}
1 reader coded the whole corpus; only the decisive
cases were regraded independently (\S\ref{sec:method}), and both readers are
authors, so the check is blind but not external. We changed the inclusion rule
after assembling the sample, and our search does not cover all relevant
English-language work. Primary-evaluation counts omit secondary studies outside
the LLM subset, both direct cases came from the targeted search, and the
disputed exclusions were not regraded. The proposed experiment tests only the next response, and
must handle missing replies, specify each move exactly, and consider whether 1
participant's interaction could affect
another's. The audit neither establishes field-wide prevalence nor shows that
successful systems require accurate explicit world models.

\section{Conclusion}
\label{sec:conclusion}
Recognizing another agent's state, obtaining a good outcome, and predicting a
response to a chosen action are different tests, and in this audit only
\NIntervForecastDirect{} primary evaluations meet the last one. Evaluations
should state which claim their scores support: when the claim concerns human
responses to proposed system actions, the forecast itself should be tested
against those responses, and the 1-step protocol is a practical place to
start.\label{sec:endmain}

\bibliographystyle{plainnat}
\bibliography{citations}

@inproceedings{yadav2026dialtom,
  title     = {{DialToM}: A Theory of Mind Benchmark for Forecasting State-Driven Dialogue Trajectories},
  author    = {Yadav, Neemesh and Achananuparp, Palakorn and Jiang, Jing and Lim, Ee-Peng},
  booktitle = {Proceedings of the 2026 Conference on Empirical Methods in Natural Language Processing (EMNLP)},
  year      = {2026},
  eprint    = {2604.20443},
  archivePrefix = {arXiv},
  url       = {https://arxiv.org/abs/2604.20443},
}

@inproceedings{zhou2024sotopia,
  title     = {{SOTOPIA}: Interactive Evaluation for Social Intelligence in Language Agents},
  author    = {Zhou, Xuhui and Zhu, Hao and Mathur, Leena and Zhang, Ruohong and Yu, Haofei and Qi, Zhengyang and Morency, Louis-Philippe and Bisk, Yonatan and Fried, Daniel and Neubig, Graham and Sap, Maarten},
  booktitle = {The Twelfth International Conference on Learning Representations (ICLR)},
  year      = {2024},
  eprint    = {2310.11667},
  archivePrefix = {arXiv},
  url       = {https://openreview.net/forum?id=mM7VurbA4r},
}

@misc{zhou2025socialworldmodels,
  title         = {Social World Models},
  author        = {Zhou, Xuhui and Liu, Jiarui and Yerukola, Akhila and Kim, Hyunwoo and Sap, Maarten},
  year          = {2025},
  eprint        = {2509.00559},
  archivePrefix = {arXiv},
  primaryClass  = {cs.CL},
  url           = {https://arxiv.org/abs/2509.00559},
}

@inproceedings{wang-etal-2024-sotopia,
  title     = {{SOTOPIA}-$\pi$: Interactive Learning of Socially Intelligent Language Agents},
  author    = {Wang, Ruiyi and Yu, Haofei and Zhang, Wenxin and Qi, Zhengyang and Sap, Maarten and Bisk, Yonatan and Neubig, Graham and Zhu, Hao},
  booktitle = {Proceedings of the 62nd Annual Meeting of the Association for Computational Linguistics (Volume 1: Long Papers)},
  year      = {2024},
  address   = {Bangkok, Thailand},
  publisher = {Association for Computational Linguistics},
  pages     = {12912--12940},
  doi       = {10.18653/v1/2024.acl-long.698},
}

@inproceedings{ys2026sotopiatom,
  title     = {{SOTOPIA-ToM}: Evaluating Privacy and Information Management in Multi-Agent Interaction with Theory of Mind},
  author    = {YS, Yashwanth and Wang, Ruichen and Zeng, Shihua and Zhou, Xuhui and Onoue, Koichi and Varadarajan, Vasudha and Sap, Maarten},
  booktitle = {Conference on Language Modeling (COLM)},
  year      = {2026},
  eprint    = {2605.02307},
  archivePrefix = {arXiv},
  url       = {https://arxiv.org/abs/2605.02307},
  note      = {arXiv:2605.02307v2},
}

@inproceedings{tsvilodub-etal-2026-emergent,
  title     = {On Emergent Social World Models {---} Evidence for Functional Integration of Theory of Mind and Pragmatic Reasoning in Language Models},
  author    = {Tsvilodub, Polina and Klumpp, Jan-Felix and Mohammadpour, Amir and Hu, Jennifer and Franke, Michael},
  booktitle = {Proceedings of the 64th Annual Meeting of the Association for Computational Linguistics (Volume 1: Long Papers)},
  month     = jul,
  year      = {2026},
  address   = {San Diego, California, United States},
  publisher = {Association for Computational Linguistics},
  pages     = {37382--37420},
  doi       = {10.18653/v1/2026.acl-long.1735},
  url       = {https://aclanthology.org/2026.acl-long.1735/},
}

@inproceedings{carroll2019overcooked,
  author        = {Carroll, Micah and Shah, Rohin and Ho, Mark K. and Griffiths, Thomas L. and Seshia, Sanjit A. and Abbeel, Pieter and Dragan, Anca},
  title     = {On the Utility of Learning about Humans for Human-{AI} Coordination},
  booktitle = {Advances in Neural Information Processing Systems},
  volume    = {32},
  year      = {2019},
  eprint    = {1910.05789},
  archivePrefix = {arXiv},
  url       = {https://arxiv.org/abs/1910.05789},
}

@article{brier1950verification,
    title   = "Verification of Forecasts Expressed in Terms of Probability",
    author  = "Brier, Glenn W.",
    journal = "Monthly Weather Review",
    volume  = "78", number = "1", year = "1950", pages = "1--3",
    doi     = "10.1175/1520-0493(1950)078<0001:VOFEIT>2.0.CO;2",
    url     = "https://journals.ametsoc.org/downloadpdf/journals/mwre/78/1/1520-0493_1950_078_0001_vofeit_2_0_co_2.pdf",
}

@article{gneiting2007strictly,
    title   = "Strictly Proper Scoring Rules, Prediction, and Estimation",
    author  = "Gneiting, Tilmann and Raftery, Adrian E.",
    journal = "Journal of the American Statistical Association",
    volume  = "102", number = "477", year = "2007", pages = "359--378",
    url     = "https://sites.stat.washington.edu/raftery/Research/PDF/Gneiting2007jasa.pdf",
}

@misc{yu2026socialworldmodels,
  title         = {Building Social World Models with {LLM}s},
  author        = {Yu, Haofei and Zhao, Yining and Lin, Guanyu and You, Jiaxuan},
  year          = {2026},
  eprint        = {2606.11482},
  archivePrefix = {arXiv},
  primaryClass  = {cs.CL},
  url           = {https://arxiv.org/abs/2606.11482},
}

@inproceedings{sclar2023symbolictom,
  title         = {Minding Language Models' (Lack of) Theory of Mind: A Plug-and-Play Multi-Character Belief Tracker},
  author        = {Sclar, Melanie and Kumar, Sachin and West, Peter and Suhr, Alane and Choi, Yejin and Tsvetkov, Yulia},
  booktitle     = {Proceedings of the 61st Annual Meeting of the Association for Computational Linguistics (Volume 1: Long Papers)},
  year          = {2023},
  month         = jul,
  pages         = {13960--13980},
  doi           = {10.18653/v1/2023.acl-long.780},
  eprint        = {2306.00924},
  archivePrefix = {arXiv},
  primaryClass  = {cs.CL},
  url           = {https://aclanthology.org/2023.acl-long.780/},
}

@inproceedings{li2023tomcollab,
  title         = {Theory of Mind for Multi-Agent Collaboration via Large Language Models},
  author        = {Li, Huao and Chong, Yu Quan and Stepputtis, Simon and Campbell, Joseph and Hughes, Dana and Lewis, Michael and Sycara, Katia},
  booktitle     = {Proceedings of the 2023 Conference on Empirical Methods in Natural Language Processing},
  year          = {2023},
  month         = dec,
  pages         = {180--192},
  doi           = {10.18653/v1/2023.emnlp-main.13},
  eprint        = {2310.10701},
  archivePrefix = {arXiv},
  primaryClass  = {cs.CL},
  url           = {https://aclanthology.org/2023.emnlp-main.13/},
}

@misc{cross2024hypothetical,
  title         = {Hypothetical Minds: Scaffolding Theory of Mind for Multi-Agent Tasks with Large Language Models},
  author        = {Cross, Logan and Xiang, Violet and Bhatia, Agam and Yamins, Daniel LK and Haber, Nick},
  year          = {2024},
  eprint        = {2407.07086},
  archivePrefix = {arXiv},
  primaryClass  = {cs.AI},
  url           = {https://arxiv.org/abs/2407.07086},
}

@inproceedings{zhu2024beliefs,
  title         = {Language Models Represent Beliefs of Self and Others},
  author        = {Zhu, Wentao and Zhang, Zhining and Wang, Yizhou},
  booktitle     = {Proceedings of the 41st International Conference on Machine Learning},
  series        = {Proceedings of Machine Learning Research},
  volume        = {235},
  pages         = {62638--62681},
  publisher     = {PMLR},
  year          = {2024},
  eprint        = {2402.18496},
  archivePrefix = {arXiv},
  primaryClass  = {cs.AI},
  url           = {https://proceedings.mlr.press/v235/zhu24o.html},
}

@inproceedings{wilf2024simtom,
  title         = {Think Twice: Perspective-Taking Improves Large Language Models' Theory-of-Mind Capabilities},
  author        = {Wilf, Alex and Lee, Sihyun Shawn and Liang, Paul Pu and Morency, Louis-Philippe},
  booktitle     = {Proceedings of the 62nd Annual Meeting of the Association for Computational Linguistics (Volume 1: Long Papers)},
  year          = {2024},
  month         = aug,
  pages         = {8292--8308},
  doi           = {10.18653/v1/2024.acl-long.451},
  eprint        = {2311.10227},
  archivePrefix = {arXiv},
  primaryClass  = {cs.AI},
  url           = {https://aclanthology.org/2024.acl-long.451/},
}

@inproceedings{hou2024timetom,
  title         = {{TimeToM}: Temporal Space is the Key to Unlocking the Door of Large Language Models' Theory-of-Mind},
  author        = {Hou, Guiyang and Zhang, Wenqi and Shen, Yongliang and Wu, Linjuan and Lu, Weiming},
  booktitle     = {Findings of the Association for Computational Linguistics: ACL 2024},
  year          = {2024},
  eprint        = {2407.01455},
  archivePrefix = {arXiv},
  primaryClass  = {cs.CL},
  url           = {https://arxiv.org/abs/2407.01455},
}

@misc{yang2025tomagent,
  title         = {Large Language Models as Theory of Mind Aware Generative Agents with Counterfactual Reflection},
  author        = {Yang, Bo and Guo, Jiaxian and Iwasawa, Yusuke and Matsuo, Yutaka},
  year          = {2025},
  eprint        = {2501.15355},
  archivePrefix = {arXiv},
  primaryClass  = {cs.CL},
  url           = {https://arxiv.org/abs/2501.15355},
}

@misc{hwang2025toma,
  title         = {Infusing Theory of Mind into Socially Intelligent {LLM} Agents},
  author        = {Hwang, EunJeong and Yin, Yuwei and Carenini, Giuseppe and West, Peter and Shwartz, Vered},
  year          = {2025},
  eprint        = {2509.22887},
  archivePrefix = {arXiv},
  primaryClass  = {cs.CL},
  url           = {https://arxiv.org/abs/2509.22887},
}

@inproceedings{kim2025thoughttracing,
  title         = {Hypothesis-Driven Theory-of-Mind Reasoning for Large Language Models},
  author        = {Kim, Hyunwoo and Sclar, Melanie and Zhi-Xuan, Tan and Ying, Lance and Levine, Sydney and Liu, Yang and Tenenbaum, Joshua B. and Choi, Yejin},
  booktitle     = {Conference on Language Modeling (COLM)},
  year          = {2025},
  eprint        = {2502.11881},
  archivePrefix = {arXiv},
  primaryClass  = {cs.AI},
  url           = {https://arxiv.org/abs/2502.11881},
}

@inproceedings{zhang2025autotom,
  title         = {{AutoToM}: Scaling Model-based Mental Inference via Automated Agent Modeling},
  author        = {Zhang, Zhining and Jin, Chuanyang and Jia, Mung Yao and Zhang, Shunchi and Shu, Tianmin},
  booktitle     = {Advances in Neural Information Processing Systems (NeurIPS)},
  year          = {2025},
  eprint        = {2502.15676},
  archivePrefix = {arXiv},
  primaryClass  = {cs.AI},
  url           = {https://arxiv.org/abs/2502.15676},
}

@inproceedings{jin2024bipalm,
  title         = {{MMToM-QA}: Multimodal Theory of Mind Question Answering},
  author        = {Jin, Chuanyang and Wu, Yutong and Cao, Jing and Xiang, Jiannan and Kuo, Yen-Ling and Hu, Zhiting and Ullman, Tomer and Torralba, Antonio and Tenenbaum, Joshua B. and Shu, Tianmin},
  booktitle     = {Proceedings of the 62nd Annual Meeting of the Association for Computational Linguistics (Volume 1: Long Papers)},
  year          = {2024},
  month         = aug,
  pages         = {16077--16102},
  doi           = {10.18653/v1/2024.acl-long.851},
  eprint        = {2401.08743},
  archivePrefix = {arXiv},
  primaryClass  = {cs.AI},
  url           = {https://aclanthology.org/2024.acl-long.851/},
}

@misc{jung2024perceptom,
  title         = {Perceptions to Beliefs: Exploring Precursory Inferences for Theory of Mind in Large Language Models},
  author        = {Jung, Chani and Kim, Dongkwan and Jin, Jiho and Kim, Jiseon and Seonwoo, Yeon and Choi, Yejin and Oh, Alice and Kim, Hyunwoo},
  year          = {2024},
  eprint        = {2407.06004},
  archivePrefix = {arXiv},
  primaryClass  = {cs.CL},
  url           = {https://arxiv.org/abs/2407.06004},
}

@inproceedings{xu2025enigmatom,
  title         = {{EnigmaToM}: Improve {LLM}s' Theory-of-Mind Reasoning Capabilities with Neural Knowledge Base of Entity States},
  author        = {Xu, Hainiu and Qi, Siya and Li, Jiazheng and Zhou, Yuxiang and Du, Jinhua and Catmur, Caroline and He, Yulan},
  booktitle     = {Findings of the Association for Computational Linguistics: ACL 2025},
  year          = {2025},
  eprint        = {2503.03340},
  archivePrefix = {arXiv},
  primaryClass  = {cs.CL},
  url           = {https://arxiv.org/abs/2503.03340},
}

@misc{lei2026rectom,
  title         = {Mind the Perspective: Let's Reason Recursively for Theory of Mind},
  author        = {Lei, Chao and Hu, Guang and Yang, Meng and Jiang, Yanbei and Lipovetzky, Nir},
  year          = {2026},
  eprint        = {2606.11724},
  archivePrefix = {arXiv},
  primaryClass  = {cs.AI},
  url           = {https://arxiv.org/abs/2606.11724},
}

@misc{qiu2023minddial,
  title         = {{MindDial}: Belief Dynamics Tracking with Theory-of-Mind Modeling for Situated Neural Dialogue Generation},
  author        = {Qiu, Shuwen and Liu, Mingdian and Li, Hengli and Zhu, Song-Chun and Zheng, Zilong},
  year          = {2023},
  eprint        = {2306.15253},
  archivePrefix = {arXiv},
  primaryClass  = {cs.CL},
  url           = {https://arxiv.org/abs/2306.15253},
}

@misc{fei2026mentis,
  title         = {Mental World Modeling},
  author        = {Fei, Hao and Zhao, Yiran},
  year          = {2026},
  eprint        = {2607.27201},
  archivePrefix = {arXiv},
  primaryClass  = {cs.CL},
  url           = {https://arxiv.org/abs/2607.27201},
}

@misc{luo2026lifelong,
  title         = {Social World Model for Lifelong Social Intelligence},
  author        = {Luo, Yu},
  year          = {2026},
  eprint        = {2606.21315},
  archivePrefix = {arXiv},
  primaryClass  = {cs.AI},
  url           = {https://arxiv.org/abs/2606.21315},
}

@misc{zhang2025swmap,
  title         = {Social World Model-Augmented Mechanism Design Policy Learning},
  author        = {Zhang, Xiaoyuan and Huang, Yizhe and Ma, Chengdong and Chen, Zhixun and Ma, Long and Du, Yali and Zhu, Song-Chun and Yang, Yaodong and Feng, Xue},
  year          = {2025},
  eprint        = {2510.19270},
  archivePrefix = {arXiv},
  primaryClass  = {cs.CY},
  url           = {https://arxiv.org/abs/2510.19270},
}

@misc{wang2026metamind,
  title         = {{MetaMind}: General and Cognitive World Models in Multi-Agent Systems by Meta-Theory of Mind},
  author        = {Wang, Lingyi and Shelim, Rashed and Saad, Walid and Ramakrishna, Naren},
  year          = {2026},
  eprint        = {2603.00808},
  archivePrefix = {arXiv},
  primaryClass  = {cs.AI},
  url           = {https://arxiv.org/abs/2603.00808},
}

@article{gmytrasiewicz2005ipomdp,
  title         = {A Framework for Sequential Planning in Multi-Agent Settings},
  author        = {Gmytrasiewicz, Piotr J. and Doshi, Prashant},
  journal       = {Journal of Artificial Intelligence Research},
  volume        = {24},
  pages         = {49--79},
  year          = {2005},
  eprint        = {1109.2135},
  archivePrefix = {arXiv},
  url           = {https://arxiv.org/abs/1109.2135},
}

@misc{he2016dron,
  title         = {Opponent Modeling in Deep Reinforcement Learning},
  author        = {He, He and Boyd-Graber, Jordan and Kwok, Kevin and Daum{\'e} III, Hal},
  year          = {2016},
  eprint        = {1609.05559},
  archivePrefix = {arXiv},
  url           = {https://arxiv.org/abs/1609.05559},
}

@misc{raileanu2018som,
  title         = {Modeling Others using Oneself in Multi-Agent Reinforcement Learning},
  author        = {Raileanu, Roberta and Denton, Emily and Szlam, Arthur and Fergus, Rob},
  year          = {2018},
  eprint        = {1802.09640},
  archivePrefix = {arXiv},
  url           = {https://arxiv.org/abs/1802.09640},
}

@misc{foerster2018lola,
  title         = {Learning with Opponent-Learning Awareness},
  author        = {Foerster, Jakob N. and Chen, Richard Y. and Al-Shedivat, Maruan and Whiteson, Shimon and Abbeel, Pieter and Mordatch, Igor},
  year          = {2018},
  eprint        = {1709.04326},
  archivePrefix = {arXiv},
  url           = {https://arxiv.org/abs/1709.04326},
}

@misc{foerster2019bad,
  title         = {{Bayesian} Action Decoder for Deep Multi-Agent Reinforcement Learning},
  author        = {Foerster, Jakob N. and Song, Francis and Hughes, Edward and Burch, Neil and Dunning, Iain and Whiteson, Shimon and Botvinick, Matthew and Bowling, Michael},
  year          = {2019},
  eprint        = {1811.01458},
  archivePrefix = {arXiv},
  url           = {https://arxiv.org/abs/1811.01458},
}

@misc{hu2020otherplay,
  title         = {{``Other-Play''} for Zero-Shot Coordination},
  author        = {Hu, Hengyuan and Lerer, Adam and Peysakhovich, Alexander and Foerster, Jakob},
  year          = {2020},
  eprint        = {2003.02979},
  archivePrefix = {arXiv},
  url           = {https://arxiv.org/abs/2003.02979},
}

@misc{hu2021obl,
  title         = {Off-Belief Learning},
  author        = {Hu, Hengyuan and Lerer, Adam and Cui, Brandon and Wu, David and Pineda, Luis and Brown, Noam and Foerster, Jakob},
  year          = {2021},
  eprint        = {2103.04000},
  archivePrefix = {arXiv},
  url           = {https://arxiv.org/abs/2103.04000},
}

@misc{zintgraf2021bayesian,
  title         = {Deep Interactive {Bayesian} Reinforcement Learning via Meta-Learning},
  author        = {Zintgraf, Luisa and Devlin, Sam and Ciosek, Kamil and Whiteson, Shimon and Hofmann, Katja},
  year          = {2021},
  eprint        = {2101.03864},
  archivePrefix = {arXiv},
  url           = {https://arxiv.org/abs/2101.03864},
}

@misc{moreno2021recursive,
  title         = {Neural Recursive Belief States in Multi-Agent Reinforcement Learning},
  author        = {Moreno, Pol and Hughes, Edward and McKee, Kevin R. and Avila Pires, Bernardo and Weber, Theophane},
  year          = {2021},
  eprint        = {2102.02274},
  archivePrefix = {arXiv},
  url           = {https://arxiv.org/abs/2102.02274},
}

@misc{papoudakis2021liam,
  title         = {Agent Modelling under Partial Observability for Deep Reinforcement Learning},
  author        = {Papoudakis, Georgios and Christianos, Filippos and Albrecht, Stefano V.},
  year          = {2021},
  eprint        = {2006.09447},
  archivePrefix = {arXiv},
  url           = {https://arxiv.org/abs/2006.09447},
}

@misc{strouse2021fcp,
  title         = {Collaborating with Humans without Human Data},
  author        = {Strouse, DJ and McKee, Kevin R. and Botvinick, Matt and Hughes, Edward and Everett, Richard},
  year          = {2021},
  eprint        = {2110.08176},
  archivePrefix = {arXiv},
  url           = {https://arxiv.org/abs/2110.08176},
}

@misc{egorov2022mamba,
  title         = {Scalable Multi-Agent Model-Based Reinforcement Learning},
  author        = {Egorov, Vladimir and Shpilman, Aleksei},
  year          = {2022},
  eprint        = {2205.15023},
  archivePrefix = {arXiv},
  url           = {https://arxiv.org/abs/2205.15023},
}

@inproceedings{sadigh2016planning,
  title     = {Planning for Autonomous Cars that Leverage Effects on Human Actions},
  author    = {Sadigh, Dorsa and Sastry, Shankar and Seshia, Sanjit A. and Dragan, Anca D.},
  booktitle = {Robotics: Science and Systems {XII}},
  year      = {2016},
  url       = {https://www.roboticsproceedings.org/rss12/p29.html},
}

@misc{liang2024gamma,
  title         = {Learning to Cooperate with Humans using Generative Agents},
  author        = {Liang, Yancheng and Chen, Daphne and Gupta, Abhishek and Du, Simon S. and Jaques, Natasha},
  year          = {2024},
  eprint        = {2411.13934},
  archivePrefix = {arXiv},
  url           = {https://arxiv.org/abs/2411.13934},
}

@misc{valevski2024gamengen,
  title         = {Diffusion Models Are Real-Time Game Engines},
  author        = {Valevski, Dani and Leviathan, Yaniv and Arar, Moab and Fruchter, Shlomi},
  year          = {2024},
  eprint        = {2408.14837},
  archivePrefix = {arXiv},
  primaryClass  = {cs.LG},
  url           = {https://arxiv.org/abs/2408.14837},
}

@misc{lewis2017deal,
  title         = {Deal or No Deal? {End-to-End} Learning for Negotiation Dialogues},
  author        = {Lewis, Mike and Yarats, Denis and Dauphin, Yann N. and Parikh, Devi and Batra, Dhruv},
  year          = {2017},
  eprint        = {1706.05125},
  archivePrefix = {arXiv},
  primaryClass  = {cs.AI},
  url           = {https://arxiv.org/abs/1706.05125},
}

@inproceedings{zhou2023supporter,
  title     = {Facilitating Multi-turn Emotional Support Conversation with Positive Emotion Elicitation: A Reinforcement Learning Approach},
  author    = {Zhou, Jinfeng and Chen, Zhuang and Wang, Bo and Huang, Minlie},
  booktitle = {Proceedings of the 61st Annual Meeting of the Association for Computational Linguistics (Volume 1: Long Papers)},
  year      = {2023},
  pages     = {1714--1729},
  publisher = {Association for Computational Linguistics},
  doi       = {10.18653/v1/2023.acl-long.96},
  url       = {https://arxiv.org/abs/2307.07994},
}

@misc{zeng2025counterfactual,
  title         = {Causal Discovery and Counterfactual Reasoning to Optimize Persuasive Dialogue Policies},
  author        = {Zeng, Donghuo and Legaspi, Roberto and Sun, Yuewen and Dong, Xinshuai and Ikeda, Kazushi and Spirtes, Peter and Zhang, Kun},
  year          = {2025},
  eprint        = {2503.16544},
  archivePrefix = {arXiv},
  primaryClass  = {cs.CL},
  url           = {https://arxiv.org/abs/2503.16544},
}

@book{pearl2009causality,
  title     = {Causality: Models, Reasoning, and Inference},
  author    = {Pearl, Judea},
  edition   = {2},
  publisher = {Cambridge University Press},
  year      = {2009},
  doi       = {10.1017/CBO9780511803161},
}

@article{robins1986new,
  title   = {A New Approach to Causal Inference in Mortality Studies with a Sustained Exposure Period---Application to Control of the Healthy Worker Survivor Effect},
  author  = {Robins, James},
  journal = {Mathematical Modelling},
  volume  = {7},
  number  = {9--12},
  pages   = {1393--1512},
  year    = {1986},
  doi     = {10.1016/0270-0255(86)90088-6},
}

@misc{zhang2026swmsurvey,
  title    = {Social World Models: A Comprehensive Survey of Theory of Mind and Social Coordination},
  author   = {Zhang, Ruoxuan and Liao, Ziqi and Zhou, Zhiyu and Jiang-Lin, Jian-Yu and Wu, Siyu and Mao, Yanbo and Wen, Bin and Xie, Hongxia and Fu, Jianlong and Yao, Meibao and Lo, Shao-Yuan and Cheng, Wen-Huang},
  year     = {2026},
  howpublished = {Survey and paper list, \url{https://github.com/avclabjlu/Awesome-Social-World-Models}},
  note     = {ResearchGate preprint, publication 395212971; no DOI or arXiv identifier as of 2026-09-14},
}

@misc{zhang2024combo,
  title         = {{COMBO}: Compositional World Models for Embodied Multi-Agent Cooperation},
  author        = {Zhang, Hongxin and Wang, Zeyuan and Lyu, Qiushi and Zhang, Zheyuan and Chen, Sunli and Shu, Tianmin and Dariush, Behzad and Lee, Kwonjoon and Du, Yilun and Gan, Chuang},
  year          = {2024},
  eprint        = {2404.10775},
  archivePrefix = {arXiv},
  primaryClass  = {cs.CV},
  url           = {https://arxiv.org/abs/2404.10775},
}

@misc{puig2023nopa,
  title         = {{NOPA}: Neurally-guided Online Probabilistic Assistance for Building Socially Intelligent Home Assistants},
  author        = {Puig, Xavier and Shu, Tianmin and Tenenbaum, Joshua B. and Torralba, Antonio},
  year          = {2023},
  eprint        = {2301.05223},
  archivePrefix = {arXiv},
  primaryClass  = {cs.RO},
  url           = {https://arxiv.org/abs/2301.05223},
}

@inproceedings{westby2023collective,
  title     = {Collective Intelligence in Human-{AI} Teams: A {Bayesian} Theory of Mind Approach},
  author    = {Westby, Samuel and Riedl, Christoph},
  booktitle = {Proceedings of the AAAI Conference on Artificial Intelligence},
  year      = {2023},
  eprint    = {2208.11660},
  archivePrefix = {arXiv},
  primaryClass  = {cs.HC},
  url       = {https://arxiv.org/abs/2208.11660},
}

@article{bhat2022clustering,
  title     = {Clustering Trust Dynamics in a Human-Robot Sequential Decision-Making Task},
  author    = {Bhat, Shreyas and Lyons, Joseph B. and Shi, Cong and Yang, X. Jessie},
  journal   = {IEEE Robotics and Automation Letters},
  year      = {2022},
  volume    = {7},
  number    = {4},
  pages     = {8815--8822},
  eprint    = {2206.01645},
  archivePrefix = {arXiv},
  primaryClass  = {cs.RO},
  doi       = {10.1109/LRA.2022.3188902},
  url       = {https://arxiv.org/abs/2206.01645},
}

@inproceedings{dragan2013legibility,
  title     = {Legibility and Predictability of Robot Motion},
  author    = {Dragan, Anca D. and Lee, Kenton C. T. and Srinivasa, Siddhartha S.},
  booktitle = {Proceedings of the 8th ACM/IEEE International Conference on Human-Robot Interaction (HRI)},
  year      = {2013},
  address   = {Tokyo, Japan},
  publisher = {IEEE},
  pages     = {301--308},
  doi       = {10.1109/HRI.2013.6483603},
  url       = {https://publications.ri.cmu.edu/storage/publications/pub_files/2013/3/legiilitypredictabilityIEEE.pdf},
}

@inproceedings{sagheb2023longterm,
  title     = {Towards Robots that Influence Humans over Long-Term Interaction},
  author    = {Sagheb, Shahabedin and Mun, Ye-Ji and Ahmadian, Neema and Christie, Benjamin A. and Bajcsy, Andrea and Driggs-Campbell, Katherine and Losey, Dylan P.},
  booktitle = {Proceedings of the 2023 IEEE International Conference on Robotics and Automation (ICRA)},
  year      = {2023},
  address   = {London, United Kingdom},
  publisher = {IEEE},
  pages     = {7490--7496},
  eprint    = {2209.10588},
  archivePrefix = {arXiv},
  primaryClass  = {cs.RO},
  doi       = {10.1109/ICRA48891.2023.10160321},
  url       = {https://arxiv.org/abs/2209.10588},
}

@article{soh2018trusttransfer,
  title     = {Multi-Task Trust Transfer for Human-Robot Interaction},
  author    = {Soh, Harold and Xie, Yaqi and Chen, Min and Hsu, David},
  journal   = {The International Journal of Robotics Research},
  year      = {2020},
  volume    = {39},
  number    = {2-3},
  pages     = {233--249},
  eprint    = {1807.01866},
  archivePrefix = {arXiv},
  primaryClass  = {cs.RO},
  doi       = {10.1177/0278364919866905},
  url       = {https://arxiv.org/abs/1807.01866},
}

@inproceedings{tian2023learning,
  title     = {Towards Modeling and Influencing the Dynamics of Human Learning},
  author    = {Tian, Ran and Tomizuka, Masayoshi and Dragan, Anca D. and Bajcsy, Andrea},
  booktitle = {Proceedings of the 2023 ACM/IEEE International Conference on Human-Robot Interaction (HRI)},
  year      = {2023},
  address   = {Stockholm, Sweden},
  publisher = {Association for Computing Machinery},
  pages     = {350--358},
  eprint    = {2301.00901},
  archivePrefix = {arXiv},
  primaryClass  = {cs.RO},
  doi       = {10.1145/3568162.3578629},
  url       = {https://arxiv.org/abs/2301.00901},
}

@article{chen2018trust,
  title     = {Trust-Aware Decision Making for Human-Robot Collaboration: Model Learning and Planning},
  author    = {Chen, Min and Nikolaidis, Stefanos and Soh, Harold and Hsu, David and Srinivasa, Siddhartha},
  journal   = {ACM Transactions on Human-Robot Interaction},
  year      = {2020},
  volume    = {9},
  number    = {2},
  pages     = {1--23},
  eprint    = {1801.04099},
  archivePrefix = {arXiv},
  primaryClass  = {cs.RO},
  doi       = {10.1145/3359616},
  url       = {https://arxiv.org/abs/1801.04099},
}

@inproceedings{kwon2020risk,
  title     = {When Humans Aren't Optimal: Robots that Collaborate with Risk-Aware Humans},
  author    = {Kwon, Minae and Biyik, Erdem and Talati, Aditi and Bhasin, Karan and Losey, Dylan P. and Sadigh, Dorsa},
  booktitle = {Proceedings of the 2020 ACM/IEEE International Conference on Human-Robot Interaction (HRI)},
  year      = {2020},
  address   = {Cambridge, United Kingdom},
  publisher = {Association for Computing Machinery},
  pages     = {43--52},
  eprint    = {2001.04377},
  archivePrefix = {arXiv},
  primaryClass  = {cs.RO},
  doi       = {10.1145/3319502.3374832},
  url       = {https://arxiv.org/abs/2001.04377},
}

@inproceedings{mandel2014ope,
  title     = {Offline Policy Evaluation Across Representations with Applications to Educational Games},
  author    = {Mandel, Travis and Liu, Yun-En and Levine, Sergey and Brunskill, Emma and Popovi{\'c}, Zoran},
  booktitle = {Proceedings of the 13th International Conference on Autonomous Agents and Multiagent Systems (AAMAS)},
  year      = {2014},
  pages     = {1077--1084},
  doi       = {10.65109/tnvs8310},
  url       = {https://grail.cs.washington.edu/wp-content/uploads/2015/08/mandel2014ope.pdf},
}

@inproceedings{nikolaidis2017game,
  title     = {Game-Theoretic Modeling of Human Adaptation in Human-Robot Collaboration},
  author    = {Nikolaidis, Stefanos and Nath, Swaprava and Procaccia, Ariel D. and Srinivasa, Siddhartha},
  booktitle = {Proceedings of the 2017 ACM/IEEE International Conference on Human-Robot Interaction (HRI)},
  year      = {2017},
  address   = {Vienna, Austria},
  publisher = {Association for Computing Machinery},
  pages     = {323--331},
  eprint    = {1701.07790},
  archivePrefix = {arXiv},
  primaryClass  = {cs.RO},
  doi       = {10.1145/2909824.3020253},
  url       = {https://arxiv.org/abs/1701.07790},
}

@article{nikolaidis2017mutual,
  title     = {Human-Robot Mutual Adaptation in Collaborative Tasks: Models and Experiments},
  author    = {Nikolaidis, Stefanos and Hsu, David and Srinivasa, Siddhartha},
  journal   = {The International Journal of Robotics Research},
  year      = {2017},
  volume    = {36},
  number    = {5-7},
  pages     = {618--634},
  doi       = {10.1177/0278364917690593},
  url       = {https://pmc.ncbi.nlm.nih.gov/articles/PMC7449140/},
}

@article{wang2024adaptiveteaching,
  title     = {A Hierarchical {Bayesian} Model of Adaptive Teaching},
  author    = {Chen, Alicia M. and Palacci, Andrew and V{\'e}lez, Natalia and Hawkins, Robert D. and Gershman, Samuel J.},
  journal   = {Cognitive Science},
  year      = {2024},
  volume    = {48},
  number    = {7},
  pages     = {e13477},
  doi       = {10.1111/cogs.13477},
  url       = {https://gershmanlab.com/pubs/Chen24.pdf},
}

@misc{zhang2026digitaltwin,
  title         = {{LLM}-Augmented Digital Twin for Policy Evaluation in Short-Video Platforms},
  author        = {Zhang, Haoting and Lin, Yunduan and He, Jinghai and Jiang, Denglin and Shen, Zuo-Jun and Zheng, Zeyu},
  year          = {2026},
  eprint        = {2603.11333},
  archivePrefix = {arXiv},
  primaryClass  = {cs.AI},
  url           = {https://arxiv.org/abs/2603.11333},
}

@inproceedings{arkhangorodsky-etal-2021-meetdot,
  title     = {{MeetDot}: Videoconferencing with Live Translation Captions},
  author    = {Arkhangorodsky, Arkady and Chu, Christopher and Fang, Scot and Huang, Yiqi and Jiang, Denglin and Nagesh, Ajay and Zhang, Boliang and Knight, Kevin},
  booktitle = {Proceedings of the 2021 Conference on Empirical Methods in Natural Language Processing: System Demonstrations},
  year      = {2021},
  pages     = {195--202},
  address   = {Online and Punta Cana, Dominican Republic},
  publisher = {Association for Computational Linguistics},
  doi       = {10.18653/v1/2021.emnlp-demo.23},
  url       = {https://aclanthology.org/2021.emnlp-demo.23/},
}

\appendix
\section{Audit Rules and Access to the Evidence}
\label{app:audit}
\label{app:coding}
\label{app:artifact}

\paragraph{Screening stages.}
The 3 stages of \S\ref{sec:method} ran on 2026-08-31 (\NSweeps{} topic-based
search sweeps), 2026-09-01 (the references of \citet{zhang2026swmsurvey} that
the sweeps had not returned) and 2026-09-09 (the targeted search for studies
that might meet the criterion).

\paragraph{The unit and primary-evaluation rule.}
1 row represents a cited work, not an experiment, dataset or checkpoint. The
released codebook calls the system whose move is being chosen the \emph{ego};
we say ``the system'' throughout the paper. The
codebook's R1 says to code what a work actually runs and scores. R2 follows the
setting its abstract and main tables emphasize; secondary studies, ablations
and appendix experiments do not determine the primary code. R5 allows 2 values
when both are primary and neither dominates. For the 4-way partition, an
assigned-action value takes precedence; otherwise the first-listed value is
used.

\NRegimeCompound{} scored works have 2 primary values: \RegimeCompoundList{}.
The terms in parentheses are the released codebook values, not additional
evaluation categories.

\paragraph{Why Qiu's human study does not set its primary code.}
\citet{qiu2023minddial} emphasizes agent--agent belief tracking and task
performance. Its 16-participant human game study is secondary (their
Section~4.1.4; interface in their Appendix~F). The primary code is therefore
task success. The separate full-text check records the human interaction, the
system-assigned action and the absence of a scored belief forecast in that
experiment. This illustrates why a primary-evaluation count and an
any-experiment count answer different questions.
Scoring a live human interaction by task success also appears outside the
audited sample: MeetDot \citep{arkhangorodsky-etal-2021-meetdot}, a
videoconferencing system with live translation captions, uses a cross-lingual
word-guessing game between its users as an extrinsic evaluation, with no
forecast of a player's response involved.

\paragraph{The admission amendment.}
The written rule admitted either a scored prediction of another agent or a
physical world-model foundation. A full-text check of the \NBoundaryChecked{}
scored, nonphysical task-success works found \NBoundaryYesForecast{} with a
scored counterpart prediction and \NBoundaryNoForecast{} without one. The
latter have unscored counterpart models or models implicit in the policy. The
rule was amended to describe this assembled sample; membership was not changed
by the amendment. The codebook records its date and rationale. The narrower
counts remove the \NBoundaryNoForecast{} cases (\NRegimeCodedNoBoundary{}
scored works), the \NFamPhysScored{} scored physical foundations
(\NRegimeCodedNoPhys{}), or both (\NStrictSubset{}). Removing both groups
leaves only nonphysical works with a scored prediction of another agent; the
original rule also admitted physical world-model foundations.

Pure benchmarks and agents that model no counterpart remain excluded. SOTOPIA
\citep{zhou2024sotopia} is not a corpus row, whereas SOTOPIA-$\pi$
\citep{wang-etal-2024-sotopia} and SOTOPIA-ToM \citep{ys2026sotopiatom} are.
Exclusion from the corpus does not establish that a work lacks a qualifying
experiment. The boundary evidence file retains an inactive proposal row for
traceability; it is not 1 of the \NBoundaryChecked{} scored cases.

\begin{figure}[htbp]
\centering
\includegraphics[width=\textwidth]{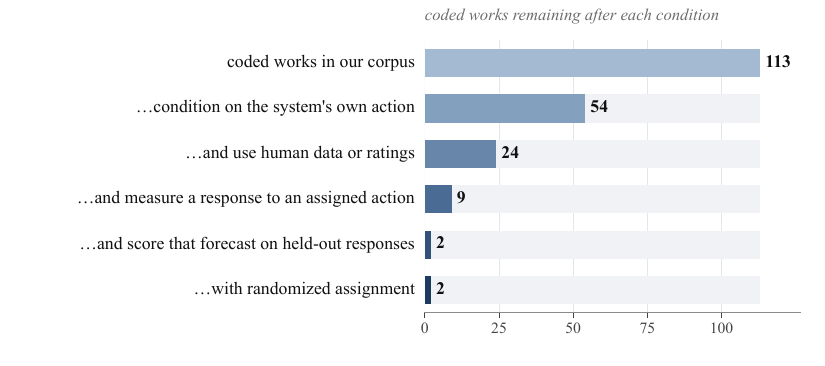}
\caption{The conditions of \S\ref{sec:question}, applied 1 at a time to the
coded works of our corpus; each bar keeps the works that pass the bar above
it, and the fifth bar is the direct forecast evaluations.}
\label{fig:contrast}
\end{figure}

\paragraph{Evidence and reproduction.}
\ifSWBcameraready
The evidence package is released, versioned and archived at \CensusDOI{}.
\else
The evidence package \CensusStatus{}.
\CensusFullLine{}
\fi
It contains the corpus
(\texttt{data/methods.csv}, \NRows{} records), the codebook and amendment
history, the screening record of all 3 passes, the study-level quotations and
source locations behind every verdict (\texttt{interventional\_evidence.csv},
1 row per assigned-action study; \texttt{llm\_family\_full\_check.csv};
\texttt{corpus\_boundary\_check.csv}; \texttt{pass3\_adjudication.csv} and
\texttt{rejected\_candidates\_grade.csv} for the targeted sweep), the recorded
source versions, and the count generator. From its root,
\texttt{python3 tools/verify\_reproduction.py} validates all \NRows{} records
and checks byte-identical regeneration of the count macros and generated
evidence tables, including the counts Figure~\ref{fig:contrast} plots; the
figure's drawing source is not included. These commands reproduce the recorded
counts; they do not establish the correctness of every coding decision.

\ifSWBcameraready
Its \texttt{README.md} records the source-version gaps.
\else
The review copy withholds identifying project metadata, and
its \texttt{README.md} says what was withheld and what was left out; corpus
codes and numerical grades are unchanged. That file also records the
source-version gaps.
\fi
The audit records a source version for
\NSourceVersionRecorded{} of the \NFamLLMChecked{} language-model papers. For
the other \NSourceVersionMissing{} it records URLs but no version identifier:
\NSourceVersionPreprint{} links point to preprints without a saved revision
number, and \NSourceVersionProceedings{} point to stable proceedings pages, where the record is fixed even though
the audit did not note a version. We do not infer which preprint revision the
original reviewer read; a later DialToM check is labeled as a new check, not as
the original audit version. The package carries the second reader's returned
sheets, their grades beside the first reader's, and the author's ruling on every
disagreement (\texttt{data/second\_reader.csv}); the separate kit for double-coding
the whole corpus is still unfilled.\ifSWBcameraready\else{} The identified repository and archive DOI will follow on
deanonymization.\fi

\section{Study-Level Evidence and Sensitivity Checks}
\label{app:evidence}

\paragraph{The \NInterventional{} assigned-action studies.}
Table~\ref{tab:allstudies} gives the assigned intervention, measured response
and forecast grade for every case. Direct means that the experiment satisfies
the criterion in \S\ref{sec:question}, including the information and
target-exclusion rules. Partial means that some forecast-related evidence
exists but does not meet that criterion. None means the human response is
measured without scoring a forecast. The source file retains quotations,
detailed inputs, metrics and the source version for each row.

\begingroup
\footnotesize
\setlength{\tabcolsep}{3pt}
\renewcommand{\arraystretch}{1.12}
\begin{longtable}{@{}P{0.18\linewidth}P{0.34\linewidth}P{\dimexpr0.48\linewidth-12pt\relax}@{}}
\caption{All \NInterventional{} assigned-action human-response studies. Grades apply to the direct-forecast criterion, not to the quality of the study as a whole. Quotations, inputs, fit timing, procedures and source versions are in \texttt{data/interventional\_evidence.csv}.}\label{tab:allstudies}\\
\toprule
Study & Assignment and human response & Forecast evidence and grade \\
\midrule
\endfirsthead
\multicolumn{3}{@{}l}{\emph{Table \thetable{} continued}}\\
\toprule
Study & Assignment and human response & Forecast evidence and grade \\
\midrule
\endhead
\midrule
\multicolumn{3}{r@{}}{\emph{Continued on next page}}\\
\endfoot
\bottomrule
\endlastfoot
\citet{dragan2013legibility} & hand-designed robot trajectories. Logged policy. \emph{Response:} observer's goal inference: time and answer. & \textbf{Partial:} ordering test (ANOVA). No forecast-error metric is reported (ordering test only); the source reports ANOVA on time/correctness score, Sec. V-C. \\[3pt]
\citet{mandel2014ope} & tutoring policy (randomized arm). Randomized. \emph{Response:} level observations; concepts completed. & \textbf{Direct:} held-out cross-validated log-likelihood. Cross-validated log-likelihood (normalized), Fig. 3 (Sec. 5, 9); reward Fig. 5. \\[3pt]
\citet{sadigh2016planning} & robot planner via best-response model. System's own policy. \emph{Response:} driver's speed, lane position, who went first. & \textbf{None:} outcome ANOVA; rollout plotted. No forecast-error metric is reported (visual comparison, Fig. 7-8); the source reports ANOVA on Raffect, Sec. V-B. \\[3pt]
\citet{nikolaidis2017mutual} & robot MOMDP policy. System's own policy. \emph{Response:} adaptation (side switch); trust rating. & \textbf{Partial:} post-hoc in-sample logistic regressions. Nagelkerke R2 36.2\%/11.9\%, 74.4\%/62.5\% classified, Sec. 8.3/9.4 (in-sample). \\[3pt]
\citet{kwon2020risk} & robot policy under 2 human models. System's own policy. \emph{Response:} tower choice; Likert ratings of the robot. & \textbf{Partial:} in-sample KL on pooled choices. Log KL divergence, Sec. 4/5.1 (online, in-sample); Sec. 5.2: time, path, Likert. \\[3pt]
\citet{soh2018trusttransfer} & randomized assignment (demo task cell). Randomized. \emph{Response:} trust rating after 2 robot demonstrations. & \textbf{Direct:} held-out NLL/MAE. Avg NLL and MAE, Tables 2-3 (Sec. 5.1-5.2); MAE DfB boxplots Figs. 10-11. \\[3pt]
\citet{bhat2022clustering} & robot MDP policy (value iteration). System's own policy. \emph{Response:} trust rating after each of 100 houses. & \textbf{Partial:} outcome-conditioned RMSE; procedure unstated. RMSE 0.1266 (SD 0.078) predicted vs. reported trust, Sec. V-A (scored set not stated); E\_RMS with 20-site training/5-site refits, Sec. V-B (clustering feature). \\[3pt]
\citet{tian2023learning} & robot teaching policy (planner). System's own policy. \emph{Response:} teleoperator's control input. & \textbf{None:} sim-only forecast error; live outcome. Theta\_H error, sims, Sec. 6.1.8; ANOVA F(1,19)=12.94 action optimality, Sec. 7. \\[3pt]
\citet{wang2024adaptiveteaching} & simulated teacher (randomized examples). Randomized. \emph{Response:} learner's decision to send a guess. & \textbf{Partial:} trend test (mixed-effects logit). No per-person forecast error is reported; the source reports mixed-effects logit bx=-35.83, bx2=-14.33, Sec. 4.3.2, Fig. 3B. \\[3pt]
\end{longtable}
\endgroup

\paragraph{Human involvement in the LLM subset.}
\label{app:llm}
Table~\ref{tab:llmfull} retains all \NFamLLMChecked{} works, including negative
codes. H means real people act in the modeled interaction or role-play; it
excludes annotators, raters of model--model conversations and people answering
benchmark questions. A means a system action is assigned in such a study. F
means a forecast is scored on human-response data, which can be recorded
dialogue; it does not itself establish decision-time eligibility. Meeting 1
condition does not imply meeting the others: a work can score forecasts from
recorded human dialogue without assigning any action to a participant. A dash
records a negative audit code, not that no person contributed to the research,
and the last column says what the full-text read found instead. All 3
requirements must hold together in 1 experiment for a direct evaluation.

\begingroup
\footnotesize
\setlength{\tabcolsep}{3pt}
\renewcommand{\arraystretch}{1.12}
\begin{longtable}{@{}P{0.28\linewidth}ccccP{\dimexpr0.72\linewidth-102pt\relax}@{}}
\caption{Full-text check of all \NFamLLMChecked{} LLM works. H: people act in the interaction or role-play. A: a system action is assigned in that study. F: a forecast is scored on human-response data. Direct requires all 3 conditions in 1 experiment with eligible inputs and scoring. Meeting 1 condition does not imply meeting the others: a work can score forecasts from recorded human dialogue without assigning any action to a participant. The last column gives the location for a positive flag and, for a negative one, what the full-text read found instead; the full quotation is in \texttt{data/llm\_family\_full\_check.csv}. Totals: H = \NFamLLMHumanStudy{}, A = \NFamLLMAssignedHuman{}, F = \NFamLLMForecastVsHuman{}, Direct = \NFamLLMAnyQualifying{}.}\label{tab:llmfull}\\
\toprule
Work & H & A & F & Direct & What was checked, and why the code is negative \\
\midrule
\endfirsthead
\multicolumn{6}{@{}l}{\emph{Table \thetable{} continued}}\\
\toprule
Work & H & A & F & Direct & What was checked, and why the code is negative \\
\midrule
\endhead
\midrule
\multicolumn{6}{r@{}}{\emph{Continued on next page}}\\
\endfoot
\bottomrule
\endlastfoot
\citet{cross2024hypothetical} & --- & --- & --- & --- & Melting Pot evaluation against scripted background bots; hypotheses scored by predictive accuracy on bot behaviour. No human participants anywhere. \\[3pt]
\citet{fei2026mentis} & --- & --- & --- & --- & Despite headline 'next action' coding and 'human-decision datasets': Menti-Bench decisions belong to fictional characters; 'target pseudo-agent and judge roles are held fixed'; the 98.5 'human reference' is people solving the benchmark, not responding to the system. \\[3pt]
\citet{hou2024timetom} & --- & --- & --- & --- & Static QA on ToMI, BigToM, FanToM; zero mentions of participants, annotators, or any human study. \\[3pt]
\citet{hwang2025toma} & --- & --- & --- & --- & SOTOPIA simulations with an LM partner and GPT-5-mini judge; CloudResearch annotators (App. B, Table 10) only validate judge reasoning. No human interacts with the agent. \\[3pt]
\citet{jin2024bipalm} & Yes & --- & --- & --- & App. B.1 Table 7 (human test set); \S{}5 Generalization \\[3pt]
\citet{jung2024perceptom} & --- & --- & --- & --- & Static QA on Percept-ToMi / Percept-FANToM; annotation by authors; no human participants. \\[3pt]
\citet{kim2025thoughttracing} & --- & --- & --- & --- & The 510 Prolific participants re-annotate BigToM stimuli (App. A); they are label providers, not interactants. All scoring is static QA on ToM benchmarks. \\[3pt]
\citet{lei2026rectom} & --- & --- & --- & --- & Static QA plus KD45 analysis; no human participants. \\[3pt]
\citet{li2023tomcollab} & --- & --- & --- & --- & LLM-only teams in a text search-and-rescue game; humans serve only as annotators of ToM-question ground truth. Limitations explicitly note human-agent teams as future work. \\[3pt]
\citet{luo2026lifelong} & --- & --- & --- & --- & ASCENT-Bench and SOTOPIA/AgentSense formal-45 runs, all LLM-simulated and LLM-judged; 'Human eval' appears only in a related-work comparison table describing SOTOPIA and Generative Agents. No participants. \\[3pt]
\citet{qiu2023minddial} & Yes & Yes & --- & --- & \S{}4.1.4 Human Evaluation, Table 4; App. F Gradio UI \\[3pt]
\citet{sclar2023symbolictom} & --- & --- & --- & --- & Only human data: 20 MTurk workers answer a ToMi template to show label ambiguity (Sec 4, footnote 3). Humans never interact with the system; all evaluation is static QA on ToMi. \\[3pt]
\citet{tsvilodub-etal-2026-emergent} & --- & --- & --- & --- & Localizer/probing analysis of 48 LMs on ToM and pragmatics benchmarks; no human participants at all. \\[3pt]
\citet{wang-etal-2024-sotopia} & --- & --- & --- & --- & Full text (arXiv 2403.08715 v3, 29 pp) read incl. App. C Human Evaluation (30 Prolific + 4 internal annotators). Humans only rate transcripts of agent-vs-GPT-3.5 dialogues on SOTOPIA-EVAL dimensions; never a conversational counterpart and nothing about them is forecast. \\[3pt]
\citet{wang2026metamind} & --- & --- & --- & --- & StarCraft Multi-Agent Challenge win rates; human-AI collaboration named only as future work. No participants. \\[3pt]
\citet{wilf2024simtom} & --- & --- & --- & --- & Static QA on ToMi/BigToM; only human involvement is 4 annotators writing oracle perspectives (Sec 6). No interaction, no forecast of a person. \\[3pt]
\citet{xu2025enigmatom} & --- & --- & --- & --- & Static QA on ToMi, HiToM, FANToM; human annotation only for entity-state training data and an appendix annotation study. No participants interact with the system. \\[3pt]
\citet{yadav2026dialtom} & --- & --- & Yes & --- & \S{}3 Prospective task; \S{}5 results; \S{}5.2 Human Baseline \\[3pt]
\citet{yang2025tomagent} & --- & --- & --- & --- & Despite headline 'next action' coding: EmpatheticDialogues/PersuasionForGood corpora only seed BDIs; the counterpart is an LLM NPC, so the 'predicted vs real utterance' comparison is against a simulated agent. Three annotators score BDI similarity; no real human responds. \\[3pt]
\citet{ys2026sotopiatom} & Yes & --- & --- & --- & App. A.13 Human Baseline, Table 7 \\[3pt]
\citet{yu2026socialworldmodels} & --- & --- & --- & --- & Forecasts prediction-market price transitions after observed real-world events; events are not assigned and prices are aggregates, not a person's response to a system action. No human participants. \\[3pt]
\citet{zhang2025autotom} & --- & --- & --- & --- & Exp. 2 (Sec 4.2) correlates posteriors with human judgments from prior cognitive-science stimuli (observational gridworld agents, no assigned action, judged not responded). App. A2 shows one human-in-the-loop debugging example; no evaluation. \\[3pt]
\citet{zhang2025swmap} & --- & --- & --- & --- & Tax, team-coordination and facility-location environments populated by RL/scripted background agents with randomly assigned latent traits; mechanisms are system-assigned but no real person responds. \\[3pt]
\citet{zhou2025socialworldmodels} & --- & --- & --- & --- & Static ToM benchmarks plus SOTOPIA-hard where 'Each episode pairs a target agent with a fixed GPT-4o partner'; humans only rate parse quality (70\% agreement with GPT-5 judge). No human interacts with the system. \\[3pt]
\citet{zhu2024beliefs} & --- & --- & --- & --- & Activation probing and steering on BigToM-derived stories; no human participants. \\[3pt]
\end{longtable}
\endgroup

\clearpage
\paragraph{The corpus-boundary check.}
Table~\ref{tab:boundaryfull} asks a different question: does the work score
\emph{any} prediction of another agent's state or action? A yes can refer to
state inference or a simulated agent. It does not establish a direct forecast
of a real human's response to an assigned system action. The
\NBoundaryChecked{} active scored cases divide into \NBoundaryYesForecast{} yes
and \NBoundaryNoForecast{} no. Their quotations and source versions remain in
\texttt{data/corpus\_boundary\_check.csv}.

\begingroup
\footnotesize
\setlength{\tabcolsep}{3pt}
\renewcommand{\arraystretch}{1.12}
\begin{longtable}{@{}P{0.19\linewidth}P{0.06\linewidth}P{0.48\linewidth}P{\dimexpr0.27\linewidth-18pt\relax}@{}}
\caption{Boundary check of the \NBoundaryChecked{} scored nonphysical task-success works: \NBoundaryYesForecast{} score some prediction of another agent, and \NBoundaryNoForecast{} do not. A yes need not concern a real person, a future response, or an assigned action. Source quotations and versions are retained in \texttt{data/corpus\_boundary\_check.csv}. The inactive proposal row is excluded.}\label{tab:boundaryfull}\\
\toprule
Work & Any score? & Predicted quantity and reported evaluation & Source location \\
\midrule
\endfirsthead
\multicolumn{4}{@{}l}{\emph{Table \thetable{} continued}}\\
\toprule
Work & Any score? & Predicted quantity and reported evaluation & Source location \\
\midrule
\endhead
\midrule
\multicolumn{4}{r@{}}{\emph{Continued on next page}}\\
\endfoot
\bottomrule
\endlastfoot
\citet{carroll2019overcooked} & Yes & The human partner's actions on the collected human-AI trajectories, using BC, H\_proxy, SP and PBT policies as predictive models of the human. \emph{Score:} Cross-entropy loss per model with s.e. over 5 seeds (Fig 7, main text); prediction accuracy (Appendix H, Fig 13). & Sec 6 'User Study', Fig 7 (main text); Appendix H 'Human-AI experiments', Fig 13 \\[3pt]
\citet{chen2018trust} & Yes & the human's action (intervene vs stay put) as a function of latent trust, and human trust dynamics (Muir-scale trust after robot success/failure). \emph{Score:} log-likelihood of the learned human behavioural model on AMT data: trust-based model -153.37 vs trust-free -156.40 (Sec. 4.3); Fig. 5 model-predicted intervention rate vs trust; Fig. 4 fitted trust transition matrices (Sec. 4.2). & Sec. 4.3 (text and Fig. 5); Sec. 4.2 Fig. 4; headline Sec. 5 Fig. 6 accumulated reward and intervention rates \\[3pt]
\citet{cross2024hypothetical} & No & Produced but not reported as scored: LLM predicts the opponent's next inventory (rock/paper/scissors) under each hypothesis about the opponent's strategy; correctness feeds an internal intrinsic-reward value V\_h used to validate hypotheses. \emph{Score:} not scored in any reported number; only ego reward is reported (Table 1 average reward and SEM; Figs. 2-5, 8, 11-16 reward). Fig. 3 plots reward around the hypothesis-validation point, not prediction accuracy. & Sec. 3 'Hypothesis Evaluation'; Sec. 4.1 Fig. 3; Table 1; Appendix E step 2 and step 4 \\[3pt]
\citet{egorov2022mamba} & No & none reported (each agent's DreamerV2-style world model predicts its own observation, reward, discount and next latent state, with attention over all agents' latents; no prediction of other agents' actions or states is scored). \emph{Score:} not scored; SMAC winning rate (Fig 4, Table 3 in appendix) and Flatland percentage of arrived trains (Fig 5, Fig 6 ablation). & Sec 5 Experiments, Fig 4 and Fig 5; Appendix Table 3 \\[3pt]
\citet{foerster2018lola} & No & none reported (LOLA-OM fits the opponent's policy parameters by maximum likelihood, Sec 4.4/Eq 4.8, but the fit is never scored). \emph{Score:} not scored; only own-colour coin percentage and points per episode (Fig 5), normalised discounted reward in IPD/IMP (Figs 2-4, Table 1), exploitability (Sec 6.3). & Sec 4.4 'LOLA with Opponent Modeling'; Sec 6.2 'Coin Game', Fig 5 \\[3pt]
\citet{foerster2019bad} & No & none about the counterpart's action/belief/goal; what is scored is the public belief over card identities in players' hands (hidden environment state), decoded from all players' actions. \emph{Score:} Fig 4c: per-card cross entropy of the belief (V0 basic, V1 self-consistent, V2 BAD/counterfactual) with the true hand during BAD play; headline is Hanabi score (Fig 4a-b, Table 2). & Sec 4 Experiments, Fig 4c and surrounding paragraph (main text) \\[3pt]
\citet{gmytrasiewicz2005ipomdp} & No & Produced but not scored: agent i's nested belief over agent j's beliefs and predicted action probabilities (belief-update traces, Figs. 5-7). \emph{Score:} not scored; the reported quantities are agent i's value functions / expected payoff for I-POMDP vs noisy POMDP (Figs. 8-10) and the multi-agent tiger game payoffs (Table 1). Belief traces are worked numerical examples with no ground-truth comparison. & Sec. 7.2 Figs. 5-7; Sec. 7.3 Figs. 8-10; Table 1 \\[3pt]
\citet{he2016dron} & No & Produced but not scored: multitask heads predict opponent's current action / buzz position (+action) and opponent type (+type) as auxiliary supervision. \emph{Score:} not scored; only ego reward is reported for the multitask variants (Table 1, Table 3 'Multitask' columns; Figs. 5, 8). No opponent-action or opponent-type prediction accuracy appears. Fig. 6a is human users' answer accuracy (data description), Fig. 6b the content model's answer accuracy, neither about the opponent. & Sec. 3.3 / Sec. 4.1 (supervision signals); Table 1; Sec. 4.2 Implementation; Table 3; Figs. 5 and 8 \\[3pt]
\citet{hu2020otherplay} & No & none reported. \emph{Score:} not scored; self-play and cross-play Hanabi scores (Fig 4, Table 1; Tables 2-4 in appendix), lever-game payoffs (Fig 3), human-AI scores (Fig 6). Fig 5 P(a\_t|a\_\{t-1\}) matrices are descriptive convention statistics from 1000 self-play episodes, not a scored prediction. & Sec 6.2 Hanabi with AI Agents, Table 1, Fig 4; Sec 6.5 Hanabi with Humans, Fig 6 \\[3pt]
\citet{hu2021obl} & No & none reported (a learned belief model predicts the active player's OWN hand to complete fictitious states, Eq 6; it is trained by supervised NLL but no belief accuracy/loss is reported). \emph{Score:} not scored; Hanabi scores for SP/XP and cross-method pairings (Table 1), toy-environment scores (Fig 3), grounded-knowledge categories of played cards (Fig 4), human-AI scores (Sec 6). & Sec 6 Hanabi Experiments, Table 1, Fig 4; Sec 4.6 / Appendix for the belief model \\[3pt]
\citet{hwang2025toma} & No & Produced but not scored: latent mental states of self and first-order beliefs about the partner (beliefs, desires, intentions, emotions, knowledge), generated between turns. \emph{Score:} not scored against any reference; mental states are selected by downstream goal-achievement score of simulated dialogue (Sec. 2.2). Reported: Sotopia-Eval Goal/Rel/Know (Tables 1-4, 8), Table 5 share of 0th- vs 1st-order reasoning, Fig. 4 distribution of mental-state dimensions, Table 10 human check of judge reasoning. & Sec. 2.2; Sec. 3.1 Metrics; Sec. 4.4 Table 5 and Fig. 4; Limitations \\[3pt]
\citet{lewis2017deal} & No & none reported as a counterpart-specific score (the seq2seq model is trained to predict tokens of both agents and is used as a forward model of the partner in rollouts, but the partner-turn prediction is never scored separately). \emph{Score:} not scored separately; Table 1 reports score/agreement/Pareto vs LIKELIHOOD agent and vs humans; Table 3 reports token perplexity and rank of the model's turns among human messages. & Sec. 3.2 and Fig. 3 (model predicts both agents' words); Sec. 5 (rollouts); Sec. 6.3 Table 3; Sec. 6.4 Table 1 \\[3pt]
\citet{li2023tomcollab} & Yes & Teammates' beliefs: first-order ToM (whether another agent knows a state/belief change) and second-order ToM (what another agent believes about the ego's knowledge). \emph{Score:} Inference accuracy (\%) vs. human-annotated ground truth from global interaction history; Table 2 (1st-order ToM: ChatGPT 41.9, GPT-4 60.0, GPT-4+Belief 80.1; 2nd-order: 11.6 / 64.3 / 69.4). & Sec. 5 'ToM inference' paragraph; Sec. 6 Table 2; question list in appendix \\[3pt]
\citet{liang2024gamma} & No & none reported (VAE decoder p(a\_t|z,tau) predicts the partner's next action and is trained on a 70/30 split, but no likelihood/accuracy of that prediction is reported). \emph{Score:} not scored; only Cooperator episode reward vs human-proxy BC agent (Fig. 4, Figs. 12-15) and vs real humans (Table 3, Fig. 5) plus Likert ratings (Fig. 6, Figs. 9-11). & Sec. 4.1 Eq. 1 (partner-action decoder); Sec. 6, Table 3, Figs. 4-6; Appendix B, F-J \\[3pt]
\citet{luo2026lifelong} & No & none reported (trajectory includes a 'mental state' field recording estimates of NPC preferences, beliefs, emotions, but it is never scored). \emph{Score:} not scored; reported metrics are ego-side: U\_final, A\_eff, S\_diag, CompletionRate, PassRate (Tables 2-6) and NPC\_final (judge-rated NPC satisfaction, an outcome not a prediction). & Sec. 3 (trajectory definition); Sec. 4.3 Core Metrics; Sec. 5 Tables 2-6 \\[3pt]
\citet{moreno2021recursive} & Yes & Opponent's private hand-crafted state (counts of rocks/papers/scissors collected) via the 0th-order belief B0, and the opponent's belief via the 1st-order belief B1; in the Tiger game P2 must guess P1's action each round. \emph{Score:} Negative log-probability belief losses L0 and L1 plotted against average reward per agent type (Fig 2c, 2d, Running with Scissors); Tiger game return where reward = 1 iff P2 guesses P1's action correctly (Fig 3). & Sec 3 Experiments, 'Running with Scissors' results paragraph, Fig 2c-d; Sec 3.1 and Fig 3 (Tiger); Appendix B (loss definition) \\[3pt]
\citet{nikolaidis2017game} & No & none reported as a scored counterpart quantity (the bounded-memory partial-adaptation model assumes the human learns the played row's payoff with probability alpha = 0.9; the model's predicted per-round reward is plotted against actual reward). \emph{Score:} not scored as a counterpart prediction; Fig. 4 left accumulated team reward (ANOVA, 42\%/52\% gains); Fig. 4 centre predicted vs actual per-round reward (a joint outcome, no error metric); text notes 73\% of participants learned vs assumed alpha = 0.9. & Sec. 8 Results and Discussion, Fig. 4 (left, centre, right) \\[3pt]
\citet{papoudakis2021liam} & Yes & The modelled (other) agent's actions and observations, reconstructed by the decoder from the controlled agent's local-observation embedding. \emph{Score:} Action reconstruction accuracy over episode time steps in double speaker-listener (>90\% after 10 steps), Fig 4b; average action prediction accuracy at step 20 with 95\% CI in 3 environments, Fig 4c. & Sec 4.4 'Model Evaluation', Fig 4b and 4c (main text) \\[3pt]
\citet{puig2023nopa} & Yes & the main agent's goal (set of goal predicates), inferred online from its actions (goal proposal network + inverse planning + particle filter); also predicted future trajectories Gamma\_k (unscored). \emph{Score:} F1-score over goal predicates of the predicted goal vs ground-truth goal, plotted over normalized task time with means and standard errors (Fig. 4b, all methods; Fig. 4c single episode); headline metric is speedup (Fig. 4a,d). & Sec. IV (metrics paragraph before Sec. V); Fig. 4(b),(c); discussed Sec. V.B \\[3pt]
\citet{qiu2023minddial} & Yes & Second-order belief b\_BinA: the speaker's estimate of the listener's belief about the mutual friend (MutualFriend task); also first-order b\_A. \emph{Score:} Precision and F1 vs. rule-based manual annotation; Table 3 (b\_BinA: LLaMA-7B 30/30, LLaMA-13B 38/33, GPT-3.5 70/67, GPT-4 76/76). & Sec. 4.1.3 'Observation III'; Table 3; Appendix A (belief annotation) \\[3pt]
\citet{raileanu2018som} & Yes & The other agent's hidden goal (z\_other), inferred online by back-propagating through the ego's own policy. \emph{Score:} Mean fraction of episodes in which the other's goal is correctly inferred (\textasciitilde{}60-80\%; \textasciitilde{}75\% on Coin), Fig 7; cumulative distribution of the step at which it is correctly inferred (\textasciitilde{}70-80\% of 1000 runs), Fig 8. & Sec 4.5 'Analyzing the goal inference', Fig 7 and Fig 8 (main text) \\[3pt]
\citet{sagheb2023longterm} & No & none reported (robot maintains a model of the human's belief b\_i over the robot's coordination reward, Eq. 6, used only to compute belief entropy in the robot's reward Eq. 7; behaviour-cloned simulated humans are built but not scored). \emph{Score:} not scored; user study 1: pass \% and lane progress over interactions (Fig. 2); simulation: human lane progress (Fig. 3); user study 2: robot reverse time (Fig. 4). & Sec. V Eq. 6-7 (unscored human-belief model); Sec. IV, VI, VII results \\[3pt]
\citet{strouse2021fcp} & No & none reported. \emph{Score:} not scored; reward with held-out agent populations (Fig 5, Table 1, Appendix C.2), reward and human preference in the human study (Fig 7), behavioural statistics such as movement frequency and pot-preference balance (Fig 8). & Sec 4.2 agent-agent evaluation, Fig 5, Table 1; Sec 5 human-agent study, Figs 7-8 \\[3pt]
\citet{wang-etal-2024-sotopia} & No & none reported. \emph{Score:} not scored; reported metrics are the trained agent's GOAL and overall Sotopia-Eval scores by GPT-4 and humans (Tables 1, 2, 5, 9), safety/engagement rates (Table 3), MMLU (Table 4). & Sec. 5 Tables 1-2; Sec. 6 Tables 3-4; Appendix Tables 5-9 \\[3pt]
\citet{wang2026metamind} & Yes & Other agents' goals (and beliefs) via inverse inference; specifically the horizon needed to identify teammates' goals. \emph{Score:} Average horizon (steps) for goal identification, reported in Fig. 5 and text: 12.3 for multi-map MetaMind vs 5.3 for single-map. No accuracy or error table. & Sec. 6.5 Scalability, Fig. 5; theory in Sec. 5 (Theorems 1-2); Appendix A.4 (inverse belief inference objective, no reported score) \\[3pt]
\citet{westby2023collective} & Yes & human teammates' beliefs and their final answers (the ToM agent shadows a player, infers the player's and neighbours' beliefs from chat messages, and predicts the human's answer choice). \emph{Score:} Human-Agent Agreement (agent's predicted answer vs human's actual answer, e.g., 75.1\% for ToM MLE) and log-likelihood with likelihood-ratio tests in Table 1; true-negative accuracy 48\% on humans' wrong answers (Results text); Fig. 3 Human-Agent Agreement curves. & Results, Model Evaluation, Table 1; Results, Analysis of Human Decision Biases; Fig. 3 \\[3pt]
\citet{ys2026sotopiatom} & No & Produced but not scored: ToM-Belief maintains per-agent AgentBelief objects (what each other agent knows / does\_not\_know / wants / thinks\_about\_me). \emph{Score:} not scored for accuracy; only downstream InfoMgmt (DA, IA, EFF, CPV) is reported (Table 2), plus a Self-Focused ablation that swaps out the belief slots (Table 3: 0.402 vs 0.329 on GPT-4o). & Sec. 4.1 metrics; Sec. 6.2 Table 3; Appendix A.9 (belief-state prompts), A.14 (ablation) \\[3pt]
\citet{zeng2025counterfactual} & No & none reported as a forecast (a GPT-2 classifier labels the dialogue strategy of observed persuadee/persuader utterances; a DDP LSTM predicts the persuadee's donation as the reward model but is never scored against ground truth). \emph{Score:} not scored as counterpart prediction; headline numbers are D3QN max/mean Q-values (Fig. 7a,b) and cumulative predicted donation \$508.78 vs \$477.62 actual (Fig. 7c); strategy-labelling classifier reports 92.3\% 5-fold accuracy (Sec. 3.2). & Sec. 3.2 (GPT-2 strategy classifier, 92.3\%); Sec. 3.4 Eq. 6 (DDP reward model); Sec. 4 Fig. 7 \\[3pt]
\citet{zhang2024combo} & Yes & other agents' intents / next actions (Intent Tracker a\_\{-i\} = IT(s\_i, G), a fine-tuned VLM predicting the cooperators' next actions from observation history). \emph{Score:} correct intent prediction rate: 75.8\% (2-agent) and 64.6\% (3-agent) TDW-Game with unseen cooperator policies (Appendix C.1); Intent Tracker test-set accuracy 79.4 (LLaVA-1.5-7B) vs 73.2 (supervised model) in Table 7. & Appendix C.1 (text, Fig. 9) and Appendix C.2 Table 7; main text Sec. 4.2 and Table 1 only score success rate / steps (incl. w/o IT ablation) \\[3pt]
\citet{zhang2025swmap} & Yes & (a) Background agents' next state/positions in response to the deployed mechanism (SWM dynamics model); (b) each background agent's latent trait (Posterior Trait Tracker). \emph{Score:} (a) state-prediction loss curves vs. MBPO/Dreamer baselines: Fig. 3b (Facility Location), Fig. 4b (Team Structure), Fig. 5d (AI-Economist); (b) trait-inference confusion matrices, Fig. 6 (Appendix B). Plotted values only; no numeric table cell. & Sec. 4.1 Fig. 3b; Sec. 4.2 Fig. 4b-c; Sec. 4.3 Fig. 5d; Appendix B Fig. 6 \\[3pt]
\citet{zhou2023supporter} & No & none reported (future emotion experts are trained to predict the positive/negative emotions of the user's next-turn utterance, Sec. 3.2 L\_emo, but their prediction accuracy is never reported). \emph{Score:} not scored; Table 2 reports PPL, Bleu, Distinct and the reward-derived cES/tES/cDC/fDC scores of the generated response; Table 3 human A/B ratings. & Sec. 3.2 (emotion experts trained on next-turn user emotion); Sec. 5.2 Table 2; Sec. 5.3 Table 3; Appendix (no accuracy) \\[3pt]
\citet{zintgraf2021bayesian} & Yes & The other agent's type/strategy, decoded from the latent belief variables (m, m\_t) by a post-hoc logistic-regression probe; the VAE action decoder also predicts other agents' future actions but that loss is not reported as a number. \emph{Score:} Test classification accuracy on 20 held-out rollouts (80 for fitting), 3 seeds, plotted over episode steps: Fig 3c 'Agent Type Prediction Accuracy'. & Sec 5.2 Treasure Hunt, Fig 3c (main text); Appendix B.2 'Classification' \\[3pt]
\end{longtable}
\endgroup

\clearpage
\paragraph{Sensitivity checks.}
\label{app:sensitivity}
Table~\ref{tab:sensitivity} varies the main coding and inclusion choices.
Removing the \NFamPhysScored{} scored physical foundations leaves
\NRegimeCodedNoPhys{} scored works; removing the \NBoundaryNoForecast{} works
without a scored counterpart prediction leaves \NRegimeCodedNoBoundary{};
applying both leaves \NStrictSubset{}. Admitting the \NMetRejectedPassThree{}
disputed candidates under their existing grades gives at most
\NDirectIfRejectedAdmitted{} of \NScorableIfRejectedAdmitted{}; that check
does not bound what unavailable evidence could establish, and the Rafferty
candidate's supplementary material could not be read, so its main-text-based
ruling remains provisional. Excluding online predict-score-update evaluations
leaves the direct count unchanged. Allowing predictors to use outcomes observed
after the action would relax the criterion, and
\citet{bhat2022clustering} would be the only additional possible case; its
evaluation procedure and scoring linkage remain unresolved, so we do not
reclassify it as direct. The original and targeted-search counts describe search
coverage rather than a limit on missed literature.

\begingroup
\footnotesize
\setlength{\tabcolsep}{3pt}
\renewcommand{\arraystretch}{1.12}
\begin{longtable}{@{}P{0.32\linewidth}P{0.35\linewidth}P{\dimexpr0.33\linewidth-12pt\relax}@{}}
\caption{Coding, eligibility, inclusion and coverage checks. Each row varies the stated decision; not every row tests robustness under the original criterion.}\label{tab:sensitivity}\\
\toprule
Decision varied & Broader count or scope & Direct evaluations / interpretation \\
\midrule
\endfirsthead
\multicolumn{3}{@{}l}{\emph{Table \thetable{} continued}}\\
\toprule
Decision varied & Broader count or scope & Direct evaluations / interpretation \\
\midrule
\endhead
\midrule
\multicolumn{3}{r@{}}{\emph{Continued on next page}}\\
\endfoot
\bottomrule
\endlastfoot
Count dialogue utterances as own-action conditioning & Conditioned works: \NEgo{} to \NEgoAlt{}; assigned-action studies: \NInterventionalAlt{}. & \NIntervForecastDirect{} (unchanged). \\[3pt]
Allow outcome-conditioned forecasts (relaxed eligibility) & The broader \NInterventional{} studies are unchanged. Inputs may now include an outcome revealed after acting. & At most \NIntervDirectOrOutcome{}; Bhat's scoring linkage is still unresolved. \\[3pt]
Exclude online predict-score-update evaluations & Fit timing changes, not the broader \NInterventional{} studies. & \NIntervForecastDirectOffline{}. \\[3pt]
Admit the \NMetRejectedPassThree{} disputed positives under their existing grades & At most \NIntervIfRejectedAdmitted{} assigned-action studies among \NScorableIfRejectedAdmitted{} scored works. & At most \NDirectIfRejectedAdmitted{} of \NScorableIfRejectedAdmitted{}; no bound on unavailable evidence. \\[3pt]
Drop the assignment requirement (different criterion) & \NIntervOneLimb{} of \NRegimeCoded{} use human data or ratings and score a predicted state or behavior. & Not a direct-forecast count or a robustness check. \\[3pt]
Exclude the \NFamPhys{} physical foundations (\NFamPhysScored{} scored) & \NIntervNoPhys{} assigned-action studies among \NRegimeCodedNoPhys{} scored works. & \NDirectNoPhys{} of \NRegimeCodedNoPhys{}. \\[3pt]
Exclude the \NBoundaryNoForecast{} unscored-counterpart cases; then also the physical foundations & \NInterventional{} of \NRegimeCodedNoBoundary{}; then \NInterventional{} of \NStrictSubset{}. & \NIntervForecastDirect{} of \NRegimeCodedNoBoundary{}; then \NStrictSubsetDirect{} of \NStrictSubset{}. \\[3pt]
Original screen versus targeted positive search (coverage) & Original: \NIntervOriginalScreen{} assigned-action study; targeted search: \NIntervFoundBySweep{} more. & Original: \NDirectOriginalScreen{}; targeted search: \NDirectFoundBySweep{} more. Not a bound on missed work. \\[3pt]
\end{longtable}
\endgroup

\clearpage
\section{Details of the Proposed Negotiation Experiment}
\label{app:protocol}

\paragraph{Outcome and payoff.}
Use 1 selected turn per participant. An observed next-turn acceptance has
$Y=1$; a reply without acceptance or an observed exit has $Y=0$. A lost or
unrecorded response is missing, not rejection. Prespecify how explicit and
ambiguous acceptance are coded. The next-turn payoff is the accepted offer's
payoff, or 0 without acceptance. Eventual negotiation payoff, if collected, is
a separate secondary outcome.

\paragraph{Baselines and uncertainty.}
The history-only and history-plus-action predictors share the training data and
model class; the candidate move is the only intended difference. The action-only
baseline predicts each move's average acceptance rate. Fit all 3 on training
participants drawn separately from the test participants, then freeze them and
log their forecasts before assignment. Both the history-only and the action-only
comparisons are primary, and a claim of success requires improvement over both.

A baseline estimated after the study is not a forecast logged before
assignment, so the 2 settings should not be mixed. A cross-fitted action-only
rate, in which each participant's rate comes from the other participants'
outcomes, may be reported separately as a retrospective diagnostic; if
bootstrapped, refit it within each replicate. Bootstrap over participants, not
turns. Choose sample size for a prespecified precision of each paired
comparison, allowing for expected missing responses. Report both paired
comparisons, per-move scores, secondary acceptance and payoff outcomes, and
participant-bootstrap intervals.

\paragraph{What missing responses can change.}
Averaging only the recorded responses can mislead if the missing replies differ
systematically from the recorded ones, and random assignment does not remove
that problem. Report the number assigned and the missing fraction for each move,
then check the best and worst results the missing responses could produce.
Because every forecast is logged before delivery, the binary outcome makes that
check simple.

For each missing reply, calculate the score difference twice: once assuming
acceptance and once assuming nonacceptance. Taking the less favorable value for
every missing reply gives the lower bound, and the more favorable value gives
the upper bound. Written out for the history-only comparison, for participant
$i$,
\[
d_i(y)=(\hat p_{0,i}-y)^2-(\hat p_{a_i,i}-y)^2,\qquad y\in\{0,1\}.
\]
Let $O$ index observed responses and $M$ missing responses, with $N=|O|+|M|$.
The mean paired difference over the full realized sample lies between
\begin{align*}
L&=\frac{1}{N}\left[\sum_{i\in O}d_i(Y_i)+\sum_{i\in M}\min\{d_i(0),d_i(1)\}\right],\\
U&=\frac{1}{N}\left[\sum_{i\in O}d_i(Y_i)+\sum_{i\in M}\max\{d_i(0),d_i(1)\}\right].
\end{align*}
The action-only comparison uses the same construction with its own logged
forecasts in place of $\hat p_{0,i}$, so each primary comparison carries its own
pair of bounds. These are missing-outcome bounds for the realized sample, not
confidence intervals. They require no missingness model, only a defined binary
target for every participant and logged forecasts. Sampling uncertainty is
separate.

\paragraph{What a positive result establishes.}
Improvement over the history-only baseline shows that knowing the move helps.
Improvement over the action-only baseline shows that the conversation helps
beyond each move's average acceptance rate. Both hold only for the delivered
moves in this task: neither reveals a person's unobserved alternative response,
establishes transfer to arbitrary messages, or shows better long-term policy
value. Randomization addresses
assignment confounding; missing outcomes, ambiguous interventions and
interference require their own treatment.

\end{document}